\documentclass[10pt,journal,compsoc]{IEEEtran}
\usepackage{graphicx}
\usepackage{amsmath,amssymb,amsfonts}
\usepackage{algorithm,algpseudocode}
\usepackage{amsthm}
\usepackage[caption=false,font=footnotesize,labelfont=sf,textfont=sf]{subfig}

\newtheorem{theorem}{Theorem}
\ifCLASSOPTIONcompsoc
  \usepackage[nocompress]{cite}
\else
  \usepackage{cite}
\fi
\ifCLASSINFOpdf
\else
\fi
\begin{document}
%
% paper title
% Titles are generally capitalized except for words such as a, an, and, as,
% at, but, by, for, in, nor, of, on, or, the, to and up, which are usually
% not capitalized unless they are the first or last word of the title.
% Linebreaks \\ can be used within to get better formatting as desired.
% Do not put math or special symbols in the title.
\title{Private and Utility Enhanced Recommendations with Local Differential Privacy and Gaussian Mixture Model}

%
%
% author names and IEEE memberships
% note positions of commas and nonbreaking spaces ( ~ ) LaTeX will not break
% a structure at a ~ so this keeps an author's name from being broken across
% two lines.
% use \thanks{} to gain access to the first footnote area
% a separate \thanks must be used for each paragraph as LaTeX2e's \thanks
% was not built to handle multiple paragraphs
%
%
%\IEEEcompsocitemizethanks is a special \thanks that produces the bulleted
% lists the Computer Society journals use for "first footnote" author
% affiliations. Use \IEEEcompsocthanksitem which works much like \item
% for each affiliation group. When not in compsoc mode,
% \IEEEcompsocitemizethanks becomes like \thanks and
% \IEEEcompsocthanksitem becomes a line break with idention. This
% facilitates dual compilation, although admittedly the differences in the
% desired content of \author between the different types of papers makes a
% one-size-fits-all approach a daunting prospect. For instance, compsoc 
% journal papers have the author affiliations above the "Manuscript
% received ..."  text while in non-compsoc journals this is reversed. Sigh.

\author{Jeyamohan Neera, Xiaomin Chen, Nauman Aslam, Kezhi Wang and Zhan Shu
    % <-this % stops a space
\IEEEcompsocitemizethanks{\IEEEcompsocthanksitem Jeyamohan Neera, Xiaomin Chen, Nauman Aslam and Kezhi Wang are with Northumbria University, UK.
% note need leading \protect in front of \\ to get a newline within \thanks as
% \\ is fragile and will error, could use \hfil\break instead.
E-mail: jeyamohan.neera@northumbria.ac.uk, xiaomin.chen@northumbria.ac.uk, nauman.aslam@northumbria.ac.uk, kezhi.wang@northumbria.ac.uk
\IEEEcompsocthanksitem Zhan Shu is with University of Alberta, Canada.
E-mail: zshu1@ualberta.ca}% <-this % stops a space
\thanks{Manuscript received ; revised  }}

% note the % following the last \IEEEmembership and also \thanks - 
% these prevent an unwanted space from occurring between the last author name
% and the end of the author line. i.e., if you had this:
% 
% \author{....lastname \thanks{...} \thanks{...} }
%                     ^------------^------------^----Do not want these spaces!
%
% a space would be appended to the last name and could cause every name on that
% line to be shifted left slightly. This is one of those "LaTeX things". For
% instance, "\textbf{A} \textbf{B}" will typeset as "A B" not "AB". To get
% "AB" then you have to do: "\textbf{A}\textbf{B}"
% \thanks is no different in this regard, so shield the last } of each \thanks
% that ends a line with a % and do not let a space in before the next \thanks.
% Spaces after \IEEEmembership other than the last one are OK (and needed) as
% you are supposed to have spaces between the names. For what it is worth,
% this is a minor point as most people would not even notice if the said evil
% space somehow managed to creep in.

% The paper headers
\markboth{Journal of \LaTeX\ Class Files,~Vol.~, No.~, }%
{Shell \MakeLowercase{\textit{et al.}}: Bare Advanced Demo of IEEEtran.cls for IEEE Computer Society Journals}
% The only time the second header will appear is for the odd numbered pages
% after the title page when using the twoside option.
% 
% *** Note that you probably will NOT want to include the author's ***
% *** name in the headers of peer review papers.                   ***
% You can use \ifCLASSOPTIONpeerreview for conditional compilation here if
% you desire.

% The publisher's ID mark at the bottom of the page is less important with
% Computer Society journal papers as those publications place the marks
% outside of the main text columns and, therefore, unlike regular IEEE
% journals, the available text space is not reduced by their presence.
% If you want to put a publisher's ID mark on the page you can do it like
% this:
%\IEEEpubid{0000--0000/00\$00.00~\copyright~2015 IEEE}
% or like this to get the Computer Society new two part style.
%\IEEEpubid{\makebox[\columnwidth]{\hfill 0000--0000/00/\$00.00~\copyright~2015 IEEE}%
%\hspace{\columnsep}\makebox[\columnwidth]{Published by the IEEE Computer Society\hfill}}
% Remember, if you use this you must call \IEEEpubidadjcol in the second
% column for its text to clear the IEEEpubid mark (Computer Society journal
% papers don't need this extra clearance.)

% use for special paper notices
%\IEEEspecialpapernotice{(Invited Paper)}

% for Computer Society papers, we must declare the abstract and index terms
% PRIOR to the title within the \IEEEtitleabstractindextext IEEEtran
% command as these need to go into the title area created by \maketitle.
% As a general rule, do not put math, special symbols or citations
% in the abstract or keywords.
\IEEEtitleabstractindextext{%
\begin{abstract}
Recommendation systems rely heavily on users’ behavioural and preferential data (e.g. ratings, likes) to produce accurate recommendations. However, users experience privacy concerns due to unethical data aggregation and analytical practices carried out by the Service Providers (SP). Local differential privacy (LDP) based perturbation mechanisms add noise to users’
data at user-side before sending it to the SP. The SP then uses the perturbed data to perform recommendations. Although LDP protects the privacy of users from SP, it causes a substantial decline in predictive accuracy. To address this issue, we propose an LDP-based Matrix Factorization (MF) with a Gaussian Mixture Model (MoG). The LDP perturbation mechanism, Bounded Laplace (BLP), regulates the effect of noise by confining the perturbed ratings to a predetermined domain. We derive a sufficient condition of the scale parameter for BLP to satisfy $\varepsilon$-LDP. At the SP, The MoG model estimates the noise added to perturbed ratings and the MF algorithm predicts missing ratings. Our proposed LDP based recommendation system improves the recommendation accuracy without violating LDP principles. The empirical evaluations carried out on three real-world datasets, i.e., Movielens, Libimseti and Jester, demonstrate that our method offers a substantial increase in predictive accuracy under strong privacy guarantee. 
\end{abstract}

% Note that keywords are not normally used for peerreview papers.
\begin{IEEEkeywords}
Data Privacy, Gaussian Mixture Model, Local Differential Privacy, Recommendation Systems
\end{IEEEkeywords}}

% make the title area
\maketitle

% To allow for easy dual compilation without having to reenter the
% abstract/keywords data, the \IEEEtitleabstractindextext text will
% not be used in maketitle, but will appear (i.e., to be "transported")
% here as \IEEEdisplaynontitleabstractindextext when compsoc mode
% is not selected <OR> if conference mode is selected - because compsoc
% conference papers position the abstract like regular (non-compsoc)
% papers do!
\IEEEdisplaynontitleabstractindextext
% \IEEEdisplaynontitleabstractindextext has no effect when using
% compsoc under a non-conference mode.

% For peer review papers, you can put extra information on the cover
% page as needed:
% \ifCLASSOPTIONpeerreview
% \begin{center} \bfseries EDICS Category: 3-BBND \end{center}
% \fi
%
% For peerreview papers, this IEEEtran command inserts a page break and
% creates the second title. It will be ignored for other modes.
\IEEEpeerreviewmaketitle

\ifCLASSOPTIONcompsoc
\IEEEraisesectionheading{\section{Introduction}\label{sec:introduction}}
\else
\section{Introduction}
\label{sec:introduction}
\fi
% Computer Society journal (but not conference!) papers do something unusual
% with the very first section heading (almost always called "Introduction").
% They place it ABOVE the main text! IEEEtran.cls does not automatically do
% this for you, but you can achieve this effect with the provided
% \IEEEraisesectionheading{} command. Note the need to keep any \label that
% is to refer to the section immediately after \section in the above as
% \IEEEraisesectionheading puts \section within a raised box.

% The very first letter is a 2 line initial drop letter followed
% by the rest of the first word in caps (small caps for compsoc).
% 
% form to use if the first word consists of a single letter:
% \IEEEPARstart{A}{demo} file is ....
% 
% form to use if you need the single drop letter followed by
% normal text (unknown if ever used by the IEEE):
% \IEEEPARstart{A}{}demo file is ....
% 
% Some journals put the first two words in caps:
% \IEEEPARstart{T}{his demo} file is ....
% 
% Here we have the typical use of a "T" for an initial drop letter
% and "HIS" in caps to complete the first word.
\IEEEPARstart{T}{he} proliferation of smartphones has boosted the usage of online shopping platforms. With more and more retailers moving online, users are feeling overwhelmed with too many options and are having trouble finding a product or service that can fulfil their expectations. Most online shopping platforms uses recommendation systems so that users can find items that could interest them. 

Collaborative Filtering (CF) is a recommendation model widely used to predict users' preference for unpurchased items. Although CF offers higher predictive accuracy, it could potentially cause privacy violations as service providers (SPs) tend to use a large amount of user data to understand and predict users' purchasing behaviour pattern. Narayanan \textit{et al.} \cite{narayanan2008} demonstrated how analyzing users' historical ratings can disclose sensitive information such as users' political preference, health related information and sometimes even their sexual orientation. Hence, it is pivotal for SPs to protect the privacy of users while providing suitable personalized recommendations.

Differential Privacy (DP) is a popular tool that can guarantee strong privacy protection even when the adversary owns a considerable amount of auxiliary information about the user \cite{dwork2014}. Most of the existing works on DP based privacy protection methods focus on protecting the privacy of user against a third-party adversary and assume that the risk of the SP causing privacy violation is minimal. Unfortunately, many SPs are apt to gather more data from users than they need and continue to procure sensitive information about users' behaviour for their own added benefits. Cambridge Analytica investigations \cite{cadwalladr2018} revealed the dangerous consequences of such harmful and unethical data aggregation practices. Not only could untrustworthy SPs violate users' privacy, but even trustworthy SPs might encounter an accidental privacy leakage as they own an enormous amount of sensitive user information. Narayanan et al. \cite{narayanan2008} deanonymized the Netflix rating data to show how trusted SPs could cause accidental data leakage.

Local Differential Privacy (LDP) \cite{Kairouz2014} has attracted much attention as it can provide strong privacy guarantee in a setting where SPs are untrustworthy. Many researchers \cite{berlioz2015,shin2018, Hua2015} have adopted LDP to protect the privacy of users in recommendation systems. In LDP-based privacy protection models, each user adds noise to his/her data locally and forward the perturbed data to the SP. As the original data never leaves the user device, users are guaranteed plausible deniability. Adopting LDP in recommendation systems cause low data utility for SPs. The predictive accuracy is comparatively higher in DP-based recommendation systems as they perturb a query output, whereas LDP-based recommendation systems add noise to the individual data point. Therefore, it is crucial to design an LDP based recommendation system which can provide strong privacy protection to users, and at the same time offer higher data utility to the SP.

Motivated by this, we propose an LDP-based recommendation system which perturbs the original ratings of a user within a predefined domain using the Bounded Lapalce (BLP) mechanism and then estimates the aggregated noise with an Mixture of Gaussian (MoG) model at the SP to enhance the data utility. The main contributions of this work are listed below:

\noindent
\begin{itemize}
    \item We introduce BLP as the input rating perturbation mechanism to increase the predictive accuracy of the LDP-based recommendation systems. To the best of our knowledge, we are the first to introduce BLP as the input data perturbation mechanism in recommendation systems and first to provide a sufficient condition for BLP to satisfy $\varepsilon$-local differential privacy in such systems. We empirically evaluate the role BLP mechanism plays in recommendation systems to enhance predictive accuracy in Section \ref{subsec:influenceBLP}.
    
    \item We introduce a noise estimation component at SP to further increase the predictive accuracy of the recommendation system. Perturbation of each user's rating leads to higher predictive error, which increases linearly with the number of users and items. We adopt Matrix Factorization (MF) with Mixture of Gaussian (MoG) to estimate the aggregated noise at SP and at the same time to predict missing ratings. This novel approach tackles low data utility issues in LDP based recommendation systems. We empirically evaluate the effect of MF with MoG in terms of achieving higher predictive accuracy in Section \ref{subse:MoGaccuracy}. We also show that the proposed LDP based recommendation model outperforms the existing LDP based recommendation models such as \cite{shin2018} and \cite{berlioz2015} in Section \ref{subsec:comparisonprediction}.
    
    \item Our approach has much lower communication cost compared to existing LDP based recommendation systems e.g.\cite{shin2018}. In our method, users only need to transmit each perturbed rating once to the SP. While on the contrary, in other systems such as \cite{shin2018}, the information exchange between a user and the SP continues for several iterations until the solution converges.
\end{itemize}

In Table \ref{tab:notationlist}, we list notations frequently used in this paper. 

\begin{table}[ht] 
\caption{Notations}
\begin{center}
 \begin{tabular}{p{70pt} p{120pt}} 
 \hline
 Notation & Meaning  \\ [0.5ex] 
 \hline
 $Pr$ & Probability \\
 $R$ & Original rating matrix   \\ 
 $R^*$  & Perturbed rating matrix  \\
$N$ & BLP noise matrix \\
 $r_{ij}$ or $r$ & Original rating \\
 $r^*_{ij}$ or $r^*$ & Perturbed rating \\
 $n_{ij}$ or $n$ & BLP noise \\
 $l$ & Minimum value in rating scale \\
 $u$ & Maximum value in rating scale \\
 $U$ & User Latent Factor Matrix \\
 $V$ & Item Latent Factor Matrix \\
 $u_i$ & rows for user $i$ in latent matrix $U$ \\
 $v_j$ & rows for item $j$ in latent matrix $V$ \\
 \hline
\end{tabular}
\end{center} 
\label{tab:notationlist}
\end{table}    

\section{Related Work} \label{sec:Relatedwork}
Generally recommendation systems use privacy protection models based on techniques such as Cryptography, obfuscation and perturbation. Cryptography based privacy protection methods can provide users with strong privacy protection. However, they usually incur a higher computational cost at user-side \cite{hong2016}. Obfuscation \cite{Param2008} and perturbation \cite{jain2011} based privacy protection approaches introduce random noise to data. Yet, the magnitude of noise added using these methods cannot be calibrated easily. DP is a popular output perturbation approach used in privacy protection models. DP based methods are proven to be a stronger solution for privacy protection in various applications compared to other peturbation methods. DP provides an information-theoretic guarantee of strong privacy protection regardless of how much knowledge adversary possesses. Unlike other perturbation approaches, in DP, calibration of noise is dependent on the sensitivity of the query and the level of privacy offered to the user.
 
Many DP based recommendation systems assume a trusted SP who collects users' ratings and releases information related to users' preferences under differential privacy guarantee. McSherry and Mironov \cite{mcsherry2009} are the first ones to integrate DP based privacy protection model with collaborative filtering-based recommendation systems. They used the Laplace mechanism to perturb the covariance matrix before predicting missing ratings. Their solution involves SP collecting original ratings from the user and then adding noise to the covariance matrix.  Yakut and Polat \cite{polat} also introduced a DP based recommendation system where user's original ratings are being stored at SP using a perturbation approach which provides uncertainty over user's actual ratings. This method also ensures that some user profiles contain fake ratings depending on the privacy budget set by the SP. 
Even though DP based recommendation models offers privacy protection to users from third party adversaries, they enable SP to collect the original ratings from users which in return causes privacy concerns. 

Hence, the attention of researchers is thus gradually shifting from DP to LDP. Many applications use LDP to deal with untrusted SPs. Google uses a randomization response mechanism RAPPOR \cite{erling2014} to collect users' Chrome usage statistics privately. However, their method is limited to simple counting queries and not extensive enough for complex aggregations. Similarly, Apple uses its own LDP mechanism to collect statistics related to emoji usage among its smartphone users \cite{apple2016}. Several works also have investigated using LDP in CF-based recommendation systems. Shen and Jin \cite{shen2014} are the first ones to investigate protecting the privacy of users from Untrusted SPs. They proposed an instance-based relaxed admissible mechanism to perturb users' private data. They aim to hide users' preference towards an item from an untrusted data aggregator. However, this method can still reveal users' preferences towards an item category. Hua \textit{et al.} \cite{Hua2015} proposed another LDP based recommendation model where the SP uses LDP based MF method to compute item profile latent factors. Subsequently, SP sends these item profiles latent factors to the users for computation of user latent factors. Each user then sends the updated item latent factors back to the SP. This method requires the users to remain online during the whole MF process. Additionally, they used an objective perturbation method for latent factor computation which adds additional communication and processing cost at the user-side.

Similarly Shin \textit{et al.} \cite{shin2018} also proposed a LDP-based recommendation model. They used a randomized response mechanism to perturb data on the user side. In their method, instead of sending the item latent factors back to the SP, each user sends back the perturbed gradient of their user latent factor matrix to the SP. This method also incurs additional processing and communication overhead to user-side as same as \cite{Hua2015}. In another work, Berlioz \textit{et al.} \cite{berlioz2015} investigated the effect of rating perturbation in different stages of the recommendation process. They evaluated the effect of input, in-process and output perturbation mechanism on the recommendation accuracy. They used a clamping method to restrict the out-of-range ratings which are perturbed using the Laplace mechanism to a pre-defined range.

\section{Preliminaries} \label{sec:Preliminaries}
\subsection{Differential Privacy}  \label{subsec:DP}
DP-based privacy protection models are relevant in settings where the SP is trusted and aggregates users' original data.  Assume two \emph{adjacent} data sets $D$ and $D'$ where $D'$ differs from $D$ by one record.  

\newtheorem{definition}{Definition}
\begin{definition}
A randomized mechanism $M$ satisfies $\varepsilon$-differential privacy if for any adjacent datasets $D$ and $D'$, and any subset $S$ of all possible outputs, we have the following inequality:
\begin{equation*}
    Pr[M(D) \in S]\leqslant{e^\varepsilon \times{Pr[M(D') \in S]}},
\end{equation*}
where $\varepsilon$ is the privacy budget.
\end{definition}
DP bounds the ability of adversary from inferring whether the input data set $D$ or $D'$ produced the given output $S$ and the privacy budget $\varepsilon$ controls the privacy loss. The smaller the value of the privacy budget $\varepsilon$, the lower the confidence the adversary has in distinguishing whether $D$ or $D'$ produced the output. Hence, DP provides a higher degree of privacy protection for lower values of privacy budget $\varepsilon$. 

\begin{definition}\label{Def:Local} Given a query $f:D \xrightarrow{} \mathbb{R}$, the sensitivity of $f$, $\Delta{f}$, can be defined as:
\begin{equation*}
    \Delta{f}=\max_{D,D'}\|f(D)-f(D')\|.
\end{equation*}
\end{definition}
The sensitivity of a function indicates how much noise is required to perturb a query result. It parametrizes the maximum difference a single record can make on the output of a query.

\subsection{Local Differential Privacy}  \label{subsec:LDP}

\begin{definition}\label{def:LDP}
A randomized mechanism $M$ satisfies $\varepsilon$-LDP if for all possible pairs of user input $x,x'$ and any subset $y$ of all possible outcomes, we have the following inequality:
\begin{equation*}
    Pr[M(x) \in y]\leqslant{e^\varepsilon \times{Pr[M(x') \in y]}}.
\end{equation*}
\end{definition}
In the LDP setting the data of each user is perturbed locally before being sent to the SP. So the SP aggregates the perturbed data instead of the original data. Therefore, even if SP possesses substantial background knowledge about the user, it cannot infer user's original data by observing the perturbed output. In this regard, LDP offers plausible deniability to users. Intuitively, LDP ensures that the SP cannot infer whether a user's input $x$ or $x'$ produce the output $y$ with confidence (controlled by the privacy budget  $\varepsilon$). 

\subsection {Laplace mechanism}  \label{subsec:LAP}
Laplace mechanism adds random noise drawn from Laplace distribution to ensure $\varepsilon$-differential privacy. We use the notation $Lap(0,b_{lap})$ to indicate that the Laplace mechanism uses a Laplace distribution with mean $0$ and scale parameter (i.e. variance) $b_{lap}$ to sample noise.

\begin{definition}
Given a query $f:D \xrightarrow{} \mathbb{R}$, the randomized mechanism $M$ satisfies $\varepsilon$-differential privacy if:
\begin{equation*}
    M(D) = f(D)+Lap(0,b_{lap}).
\end{equation*}
\end{definition}

The scale parameter $b_{lap}$ controls the width of Laplace distribution. If $\Delta{f}$ is the sensitivity of the query $f$ and $\varepsilon$ is the privacy budget, then the scale parameter $b_{lap}$ of Laplace distribution can be determined as:
\begin{equation*} 
    b_{lap}=\frac{\Delta{f}}{\varepsilon}.
\end{equation*}
Hence, the width of the Laplace distribution is dependent on sensitivity $\Delta f$ and privacy budget $\varepsilon$.

\subsection{Matrix Factorization}  \label{subsec:MF}
Matrix Factorization algorithm is the state-of-the-art technology used in CF-based recommendation systems. Many E-commerce platforms prefer MF over other methods due to it's higher predictive accuracy and computational scalability. Input for MF algorithm is the rating matrix $R$ which contains ratings of $m$ users over $n$ items. Each element $r_{ij}$ in the rating matrix indicates the rating of user $i \in \{1,2,\cdots,m\}$ on item $j \in \{1,2,\cdots,n\}$. Generally, rating matrices are sparse data sets as users tend to rate only a small group of items. MF algorithm predicts the missing rating by modelling the interactions between users and items as the inner product of latent factor spaces.

MF algorithm factorizes the given rating matrix $R$ into two latent matrices: $U$ (user latent factor matrix) and $V$ (item latent factor matrix). MF obtains the user and item latent matrices by minimizing the squared error for all known ratings in the rating matrix. 
\begin{equation}\label{eq:Objectivefunction}
    \min_{U,V} \sum_{r_{ij} \in R} [r_{ij}-u_i^Tv_j]^2.
\end{equation}
In Eq. (\ref{eq:Objectivefunction}), $u_i$ represents the relationship between user $i$ and the latent factors in the user latent matrix $U$. Similarly, $v_j$ represents the relationship between item $j$ and the latent factors in the item latent matrix $V$. The non-convex optimization problem given by Eq. (\ref{eq:Objectivefunction}) is solved using either stochastic gradient descent (SGD) or alternating least squares (ALS) method. After obtaining the latent matrices, MF predicts the missing rating of a user on an item using the dot product of the corresponding user and item latent column vectors:
\begin{equation*}
    \hat{r_{ij}} = u_i^Tv_j.
\end{equation*}

\section{Local Differential Privacy Recommendation with BLP and MoG}  \label{sec:LDPSystem}
\begin{figure}[htp]
\centering
\includegraphics[width=0.5\textwidth]{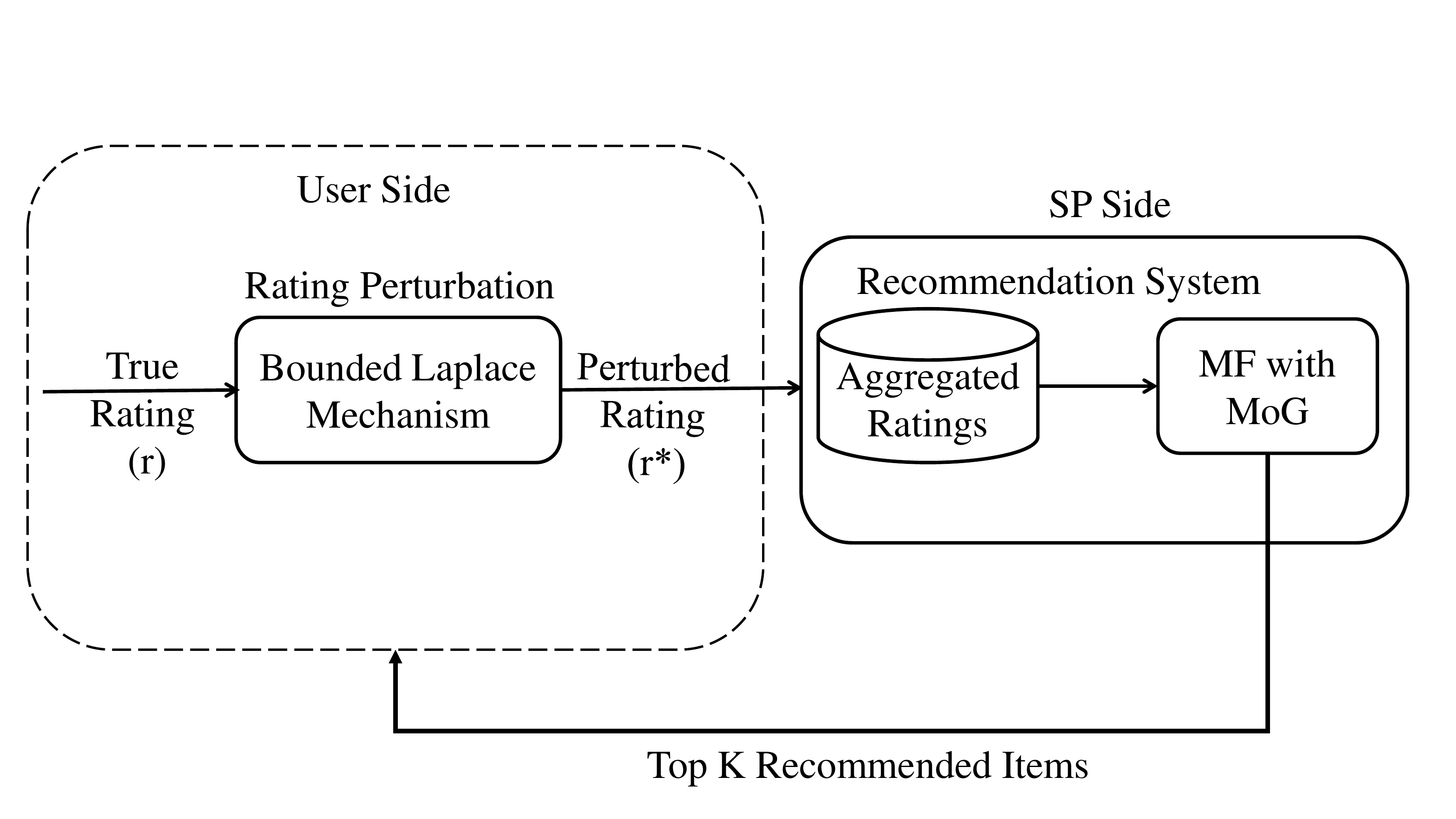}
\caption{LDP based MF recommendation with MoG}
\label{fig:system Diagram}
\end{figure}

Our proposed recommendation model is applicable in a setting where the users are cautious about sharing sensitive information with an untrustworthy SP.  Fig. \ref{fig:system Diagram} illustrates the proposed recommendation system. An LDP mechanism, BLP, perturbs the true ratings of a user before sending to the SP. Hence, the SP can only aggregate perturbed ratings from the users. At the SP,  MF with MoG model estimates the noise added to the ratings and perform missing rating prediction. Post-processing property of LDP implies that further processing a perturbed output of a $\varepsilon$-differentially private mechanisms does not cause any adverse effects on privacy protection \cite{dwork2014}. Since LDP mechanisms are immune to post-processing, estimating noise at the SP-side does not cause any additional privacy risk to users. We will describe each component of the system in detail in this section. 

\subsection{LDP Rating Perturbation} 

\subsubsection{Bounded Laplace Mechanism}  \label{subsec:BLP}
As discussed in section 3, the Laplace mechanism achieves $\varepsilon$-differential privacy by sampling random noise within the range of $-\infty$ to $\infty$. The perturbed output thus falls within the domain of $-\infty$ to $\infty$. For example, if we add noise to a user rating using the Laplace mechanism, it might produce a negative result as an output. Although this negative output holds no physical meaning in terms of the rating scale, it is still a valid output of the Laplace mechanism. Such inconsistent perturbed ratings have an immense effect on the predictive accuracy of MF based recommendation systems. 

We use BLP as input perturbation mechanism to increase the predictive accuracy of LDP based recommendation systems. The BLP mechanism ignores off-limit values and samples noise for a given input rating continuously until a perturbed rating falls within the predefined output domain. Given an input rating $r$, the BLP mechanism continuously samples noise from a Laplace distribution until the perturbed rating $r^*$ falls within the predefined output domain, i.e. $l \leq r^* \leq u$, where $l$ is the minimum and, $u$ is the maximum value of the given rating scale. 

Bounded Laplace mechanism can be defined using the probability density function (pdf) as below \cite{holohan2018}: 

\begin{definition}\label{def:BLP}
Given a domain interval of $(l,u)$, input $r \in [l,u]$ and the scale parameter $b > 0$, the Bounded Laplace mechanism $W$,  is given by the conditional probability density function :
\begin{equation*}
    f_{W}(r^*) = \begin{cases}
    \frac{1}{C(r)}\frac{1}{2b}e^{-\frac{|r^*-r|}{b}} ,& \text{if } r^* \in [l,u],\\
    0,              & \text{if } r^* \notin [l,u],
\end{cases}
\end{equation*}
where $C(r) = \int_{l}^{u}\frac{1}{2b}e^{-\frac{|r^*-r|}{b}}dr^*$ is a normalization factor dependent on input $r$.
\end{definition}

\newtheorem{lem}{Lemma}
\begin{lem}\label{lem:normconst}
The normalization factor $C(r)$ is given  by:
\begin{equation*}
    C(r)=1-\frac{1}{2}\bigg(e^{-\frac{r-l}{b}}+e^{-\frac{u-r}{b}} \bigg).
\end{equation*}
\begin{proof}

\begin{equation*} 
\begin{split}
    C(r) &= \int_{l}^{u}\frac{1}{2b}e^{-\frac{|r^*-r|}{b}}dr^* \\
    &=\int_{l}^{r}\frac{1}{2b}e^{\frac{r^*-r}{b}}dr^*+\int_{r}^{u}\frac{1}{2b}e^{-\frac{r^*-r}{b}}dr^* \\
    &=\frac{b}{2b} \left[e^{\frac{r^*-r}{b}} \right]_l^r + \frac{b}{2b} \left[-e^{-\frac{r^*-r}{b}} \right]_r^u\\
    &=1-\frac{1}{2}\bigg(e^{-\frac{r-l}{b}}+e^{-\frac{u-r}{b}} \bigg).
\end{split}
\end{equation*}
\end{proof}
\end{lem}
Assume that $r$ and $r'$ are a pair of possible inputs to a randomized mechanism and  $r'=r+z$. We will define $F(r,z)$ as:
\begin{equation*}
   F(r,z)= \frac{C({r+z})}{C(r)}e^\frac{\mid r'-r\mid}{b}.
\end{equation*}
\begin{lem} \label{lem:CRproperty}
 Let  $0  \leq z \leq \Delta f$, then,
\begin{equation*}
    \max_{\substack{r,r' \in [l,u] \\ 0  \leq z \leq \Delta f}} F(r,z)= \frac{C({l+\Delta f})}{C(l)}e^{\frac{\Delta f}{b}}.
\end{equation*}

\begin{proof}
The full proof is given in Appendix A.
\end{proof}
\end{lem}

\noindent We will define $\Delta C$ for later use:
\begin{equation*}
    \Delta C = \frac{C(l+\Delta f)}{C(l)}.
\end{equation*}

 \begin{theorem} \label{theo:BLPscale}
 When scale parameter $b \geq \frac{\Delta f}{\varepsilon-log \Delta C}$, it is sufficient to show that the Bounded Laplace mechanism $W$ satisfies $\varepsilon$-local differential privacy 
 \end{theorem}
 
 \begin{proof}
 Assume that $r$ and $r'$ are a pair of possible inputs to a Bounded Laplace mechanism and  $r'=r+z$. Let  $0  \leq z \leq \Delta f$. $r^*$ represents a possible perturbed output produced by BLP mechanism.  Given the domain of the perturbed output is $[l,u]$, we can note that,
 \begin{equation*}
     Pr(W(r) \in [l,u] )= \frac{1}{C(r)}Pr(M(r) \in [l,u]),
 \end{equation*}
 where $M$ represents the Laplace mechanism.  
 
 We aim to find a condition under which $W$ satisfies $\varepsilon$-local differential privacy. Based on the LDP definition, we can note that,
 \begin{equation*}
 \begin{split}
      Pr(W(r) \in [l,u]) &\leq e^\varepsilon Pr(W(r') \in [l,u]), \\
      \frac{1}{C(r)} Pr(M(r) \in [l,u])&\leq e^\varepsilon \frac{1}{C(r')} Pr(M(r') \in [l,u]).
 \end{split}
\end{equation*}
Given that $Pr(M(r) \in [l,u]) = \int_{l}^{u}\frac{1}{2b}e^{-\frac{|r^*-r|}{b}}dr^* $, we have,
\begin{equation}\label{eq:theorem1eq1}
    \frac{1}{C(r)} \int_{l}^{u}\frac{e^{-\frac{|r^*-r|}{b}}}{2b}dr^* \leq e^{\varepsilon}  \frac{1}{C(r')} \int_{l}^{u}\frac{e^{-\frac{|r^*-r'|}{b}}}{2b}dr^*.
\end{equation}

A lower bound for $e^{\varepsilon}  \frac{1}{C(r')} \int_{l}^{u}\frac{e^{-\frac{|r^*-r'|}{b}}}{2b}dr^*$ can be obtained using the triangle inequality, i.e.
\begin{equation*} \label{eq:trainglefurther}
\begin{split}
\mid r^*-r'\mid &\leq \mid r^*-r \mid + \mid r'-r \mid ,\\
     e^\varepsilon\frac{1}{C(r')} \int_{l}^{u}\frac{e^{-\frac{|r^*-r'|}{b}}}{2b}dr^* &\geq e^{\varepsilon}  \frac{1}{C(r')} \int_{l}^{u}\frac{e^{-\frac{|r^*-r|+\mid r'-r \mid}{b}}}{2b}dr^*,\\
     e^\varepsilon\frac{1}{C(r')} \int_{l}^{u}\frac{e^{-\frac{|r^*-r'|}{b}}}{2b}dr^* &\geq e^{\varepsilon-\frac{\mid r'-r \mid}{b}}  \frac{1}{C(r')} \int_{l}^{u}\frac{e^{-\frac{|r^*-r|}{b}}}{2b}dr^*.
\end{split}
\end{equation*}

To ensure Eq. (\ref{eq:theorem1eq1}) hold, it is sufficient to show that:
\begin{equation} \label{eq:12rewritten}
    \frac{1}{C(r)} \int_{l}^{u}\frac{1}{2b}e^{-\frac{|r^*-r|}{b}}dr^* \leq e^{\varepsilon-\frac{\mid r'-r\mid}{b}} \frac{1}{C(r')} \int_{l}^{u}\frac{1}{2b}e^{-\frac{|r^*-r|}{b}}dr^*.
\end{equation}
The inequality given by Eq. (\ref{eq:12rewritten}) can be further reduced as,
\begin{equation*}
    \frac{C(r)}{C(r')} e^{\varepsilon-\frac{\mid r'-r\mid}{b}}  \geq 1 .
\end{equation*}

\noindent From Lemma \ref{lem:CRproperty} we can note that,
\begin{equation*}
    \frac{C(r')}{C(r)} e^{\frac{\mid r'-r\mid}{b}} \leq \Delta C e^{\frac{\Delta f}{b}} , 
\end{equation*}

\noindent Equivalently,
\begin{equation*}
    \frac{C(r)}{C(r')} e^{\varepsilon-\frac{\mid r'-r\mid}{b}} \geq \frac{1}{\Delta C} e^{\varepsilon-\frac{\Delta f}{b}}.
\end{equation*}

\noindent We find a lower bound for $\frac{C(r)}{C(r')} e^{\varepsilon-\frac{\mid r'-r\mid}{b}}$ and proceed to find a condition for Eq. (\ref{eq:12rewritten}) to hold. \\

\noindent To make Eq. (\ref{eq:12rewritten}) hold, it is sufficient to show that,
\begin{equation*}
    1 \leq e^{\varepsilon-\frac{\Delta f}{b}} \frac{1}{\Delta C}.
\end{equation*}

\noindent or equivalently, 
\begin{equation*}
    b \geq\frac{\Delta f}{\varepsilon - \log (\Delta C)}.
\end{equation*}
\end{proof}

Theorem \ref{theo:BLPscale} provides the scale parameter for BLP to satisfy $\varepsilon$-local differential privacy. It also demonstrates that BLP cannot satisfy $\varepsilon$-local differential privacy when inheriting the scale parameter from the Laplace mechanism. In our recommendation system, we use BLP as input rating perturbation mechanism. Input perturbation mechanism calibrates the magnitude of noise added to original ratings according to the sensitivity given by $\Delta f =u-l$. \\

\noindent We defined $\Delta C$ as:
\begin{equation*}
    \begin{split}
        \Delta C &= \frac{C(l+\Delta f)}{C(l)} \\
        &= \frac{1-\frac{1}{2}(e^{-\frac{\Delta f}{b}}+e^{-\frac{u-\Delta f-l}{b}})}{1-\frac{1}{2}(1+e^{-\frac{u-l}{b}})}.
    \end{split}
\end{equation*}
When $\Delta f=u-l$, 
\begin{equation*}
\begin{split}
    \Delta C &= \frac{1-\frac{1}{2}(1+e^{-\frac{(u-l)}{b}})}{1-\frac{1}{2}(1+e^{-\frac{(u-l)}{b}})} \\
    &= 1.
\end{split}
\end{equation*}

\noindent Thus $\log \Delta C=0$.

Therefore we can conclude that a sufficient condition needed for BLP mechanism to satisfy $\varepsilon$-local differential privacy in our recommendation system can be given by:
\begin{equation*}
    b \geq \frac{u-l}{\varepsilon}.
\end{equation*}

Algorithm 1 details the stages involved in generating a perturbed rating using BLP mechanism.

\begin{algorithm}
\caption{BLP Mechanism for Noise Sampling}
\begin{algorithmic}[1]
\State {{\textbf{Input to the Mechanism: Original Rating ($r$)}}}
\State{\textbf{Output of the Mechanism: Perturbed Rating ($r^*$)}}
\State A noise value is generated from the Laplace distribution with mean 0 and variance of $b$ - $Lap(0,b)$
\State Add noise to original rating to obtain perturbed rating: $r^* = r+Lap(0,b)$
\State {\textbf{If} ({$r^* \in (l,u)$}):}
    \State {$\> \>$ Perturbed rating is set to $r^*$}
\State {\textbf{else}}
    \State $\> \>$ repeat Step 3 until ($r^* \in (l,u)$)
 \State {\textbf{Return}} Perturbed rating to SP
 \end{algorithmic} 
\end{algorithm}

\subsubsection{BLP Noise Distribution} \label{subsec:noisedistribution}
The noise distribution of BLP mechanism can be theoretically derived for any given dataset of true input ratings. Consider a discrete rating system containing $h$ evenly distributed discrete ranks with step size $c$. The rank set is denoted by  $\mathcal{Q}=\{Q_1,..,Q_{h-1},Q_h\}$, and $\mid Q_{i+1}-Q_i \mid=c,  1\leq i \leq h-1$. Let $r$ be a true rating, its corresponding perturbed rating is $r^*=r+n$, where $n$ is the random noise drawn by BLP mechanism. Since $r^*$ can only take values in the set $\mathcal{Q}$, i.e. $Q_1\leq {r^*} \leq Q_h$, we have the noise range for input rating $r$ as $Q_1-r \leq n \leq Q_h-r$. Define the probability $\theta_r$ as:
\begin{equation*}
    \theta_r = Pr\big(Q_1-r \leq n < Q_h-r\big),
\end{equation*}
$\theta_r$ represents the probability that the noise variable $n$ falls into the interval $(Q_1-r,Q_h-r)$ given an input rating $r$. The input rating $r$ takes values in a finite set $\mathcal{Q}$. $\theta_r$ can thus be expanded as
\begin{equation*}
    \theta_r = \sum_ {Q_i \in \mathcal{Q}} Pr(r=Q_i) Pr\big (Q_1-Q_i\leq n < Q_h-Q_i| r=Q_i\big)
\end{equation*}
where $Pr(r=Q_i)$ is the probability that the input rating equals to $Q_i$, and $Pr\big(Q_1-Q_i\leq n < Q_h-Q_i| r=Q_i\big)$ is the conditional probability that the BLP noise lies within the interval $(Q_1-Q_i \leq n < Q_h-Q_i)$ under the condition that the input rating is $Q_i$. Note that not all the input ratings in $\mathcal{Q}$ leads to the noise $n$ falling within this particular range. When the perturbed rating $r^* \notin \mathcal{Q}$, the conditional probability $Pr\big(Q_1-Q_i \leq n < Q_h-Q_i| r=Q_i\big)$ yields 0. 

The noise added by the BLP mechanism over all possible input ratings in $\mathcal{Q}$ is a random variable ranging within $(Q_1-Q_h \leq n < Q_h-Q_1)$. We will then divide the range into equal intervals. The length of each interval is the rank step size of $c$. The probability that the noise variable lies within each interval given as follow:
 \begin{equation}\label{NoiseDistribution1}
 \begin{split}
     & Pr\bigg(Q_1-Q_{h-t} \leq n < Q_1-Q_{h-t-1}\bigg) = \sum_{i=h-t}^{h} Pr(r=Q_i)\\
        & \cdot Pr\bigg(Q_1-Q_{h-t} \leq n < Q_1-Q_{h-t-1}| r=Q_i\bigg)\\
& \forall t=0, \cdots , h-2. 
  \end{split}
\end{equation}
and 
\begin{equation}\label{NoiseDistribution2}
 \begin{split}
    & Pr\bigg(Q_h-Q_{t+1} \leq n < Q_h-Q_t\bigg) =  \sum_{i=1}^{t} Pr(r=Q_i)\\
    &\cdot Pr\bigg(Q_h-Q_{t+1} \leq n < Q_h-Q_t | r=Q_i\bigg)\\
 & \forall t=2, \cdots , h-1.
 \end{split}
\end{equation}

\noindent The conditional probability is given by 
\begin{equation*}
\begin{split}
 &Pr\bigg(n \in [Q_1-Q_{h-t}, Q_1-Q_{h-t-1})| r=Q_i\bigg)\\ &=\int_{Q_1-Q_{h-t}}^{Q_1-Q_{h-t-1}}\frac{1}{C_r}\frac{1}{2b}e^{-\frac{|r^*-x|}{b}}dr^* \\
 &=\frac{1}{C_r}\frac{1}{2b}(e^{Q_1-Q_{h-t-1)}}-e^{Q_1-Q_{h-t}}) \\
& \text{for} \quad t=0, \cdots , h-2. 
 \end{split}
\end{equation*}

\noindent and
\begin{equation*}
\begin{split}
 &Pr\bigg(n \in [Q_h-Q_{t+1}, Q_h-Q_t)| r=Q_i\bigg)\\ &=\int_{Q_h-Q_{t+1}}^{Q_h-Q_t}\frac{1}{C_r}\frac{1}{2b}e^{-\frac{|r^*-x|}{b}}dr^* \\
 &=\frac{1}{C_r}\frac{1}{2b}(e^{Q_h-Q_t}-e^{Q_h-Q_{t+1}}) \\
 & \text{for} \quad t=2, \cdots , h-1.
 \end{split}
\end{equation*}
where $C_r = 1-\frac{1}{2}\bigg(e^{-\frac{Q_i-Q_1}{b}}+e^{-\frac{Q_h-Q_i}{b}} \bigg)$.

\subsection{Noise Estimation with Mixture of Gaussians} \label{subsec:noiseestimation}
The MoG model is widely used to approximate probability distributions with no closed-form expression. In image processing this model is being used for the purpose of image segmentation \cite{Vidal2007}, image compression \cite{Turk1991} and background subtraction \cite{Meng2013}. We propose an MoG with MF recommendation model to estimate the noise added to the true ratings and perform missing rating prediction. Since a multivariate Gaussian distribution can model the uncertainty of a noise data point, MoG is a good solution for noise estimation. 

Since we add BLP noise to each true rating in the true rating matrix $R$, the perturbed rating matrix $R^*$ can thus be given by:
\begin{equation*}
    R^*=R+N.
\end{equation*}
We aim to find a mixture of $K$ Gaussian components to best represent the noise distribution. We assume that each noise data point $n_{ij}$ in $N$ is drawn from a Gaussian distribution $\mathcal{N}(n_{ij} \mid 0,\sigma_k^2)$ where $\sigma_k$ is the standard deviation of the $k$-th Gaussian component. The mixture of $K$ Gaussian components representing the noise data point $n_{ij}$ can thus be given by:
\begin{equation*}
    p(n_{ij} \mid \Pi, \Sigma) \sim \sum_{k=1}^K \pi_k \mathcal{N}(n_{ij} \mid 0,\sigma_k^2),
\end{equation*}
in which $\pi_k$ ($\sum_{k=1}^K \pi_k=1$) is the mixture proportion representing the probability that $n_{ij}$ is drawn from the $k$-th mixture component. $\Pi=(\pi_1,\pi_2,....\pi_k)$ and $\Sigma=(\sigma_1,\sigma_2,....\sigma_k)$. As given in preliminaries, each known rating in original rating matrix $R$ can be approximated using MF as:
\begin{equation*}
    r_{ij}= (u_i^T)v_j.
\end{equation*}
Hence each rating $r^*_{ij}$ in the perturbed rating matrix can be given by:
\begin{equation*}
    r^*_{ij}=r_{ij}+n_{ij} = (u_i^T)v_j+n_{ij}.
\end{equation*}
 Subsequently, the probability distribution of perturbed rating $r^*_{ij}$ can then be given by:
\begin{equation*}
    p(r^*_{ij}\mid u_i,v_j,\Pi,\Sigma) =\sum_{k=1}^{K}\pi_k\mathcal{N}(r^*_{ij}\mid (u_i^T)v_j,\sigma_k^2).
\end{equation*}
The likelihood of $R^*$ can thus be given by:
\begin{equation*}
    p(R^* \mid V,U,\Sigma,\Pi)=\prod_{i,j \in \Omega} \sum_{k=1}^{K}\pi_k\mathcal{N}(r^*_{ij}\mid (u_i^T)v_j,\sigma_k^2),
\end{equation*}
where $\Omega$ represents the set of non-missing data points in perturbed rating matrix $R^*$. Given the likelihood, next, we derive the maximum likelihood estimates of the model parameters $V, U,\Sigma$ and $\Pi$ for the perturbed rating matrix $R^*$, i.e.:
\begin{equation} \label{eq:MoGLike}
\begin{split}
    \max_{V,U, \Sigma, \Pi}\ & \mathcal{L}(R^* \mid V,U,\Sigma, \Pi) \\
    = &\sum_{i,j \in \Omega,}\log \sum_{k=1}^{K} \bigg( \pi_k\mathcal{N}(r^*_{ij} \mid (u_i^T)v_j,\sigma_k^2) \bigg).
\end{split}
\end{equation}
The log-likelihood can be simplified as \cite{demp1977}:
\begin{equation}\label{eq:maxlikerewritten}
    \max_{U,V, \Pi,\Sigma} \sum_{i,j \in \Omega} \sum_{k=1}^{K} \gamma_{ijk}\bigg(\log \pi_k -\log \sqrt{2\pi}\sigma_k-\frac{(r^*_{ij}-(u_i^T)v_j)^2}{2\sigma_k^2}\bigg).
\end{equation}
\subsection{Expectation Maximization for MoG}
We use Expectation-Maximization (EM) method \cite{demp1977} to evaluate and compute model parameters $V, U,\Sigma$ and $\Pi$ to maximize the likelihood function given by Eq. (\ref{eq:MoGLike}). The EM is an iterative algorithm that can be summarized as follow:
\begin{itemize}
    \item Initialize the model parameters 
    \item Evaluate the initial value of log-likelihood 
    \item Expectation (E-Step) : Evaluate the posterior responsibilities using the current model parameters
    \item Maximization (M-Step) : Re-estimate the model parameters using the current posterior responsibilities
\end{itemize}
EM algorithm updates the parameters and alternates E-step and M-step until convergence. The standard EM algorithm estimates the mean of each cluster at every iteration.  Whereas in our system, the clusters share the same parameters $U$ and $V$.

 At first, we randomly initialize the model parameters $V, U,\Sigma$ and $\Pi$ to estimate posterior responsibilities of K Gaussian components. In E-step we estimate the posterior responsibility for each noise point $n_{ij}$ using the current model parameters $V, U,\Sigma$ and $\Pi$ as:
\begin{equation}\label{eq:posteriodresp}
\gamma_{ijk} = \frac{\pi_k\mathcal{N}(r^*_{ij} \mid (u_i^T)v_j,\sigma_k^2)}{\sum_{k=1}^{K} \pi_k\mathcal{N}(r^*_{ij} \mid (u_i^T)v_j,\sigma_k^2)}.
\end{equation}
The posterior responsibility reflects the probability that $k$-th Gaussian component produces the noise point $n_{ij}$. Then in M-step, we re-estimate each model parameter $V, U,\Sigma$ and $\Pi$ based on the posterior responsibilities $\gamma_{ijk}$ from E-step. 
We first update $\Pi$ and $\Sigma$:
\begin{equation*}
   {S_k}^{(x+1)}= \sum_{i,j \in \Omega} {\gamma_{ijk}}^{(x)},
\end{equation*}
\begin{equation*}
    {\pi_k}^{(x+1)}=\frac{{S_k}^{(x+1)}}{S},
\end{equation*}
\begin{equation}\label{eq:paramupdate}
   \sigma_k^2=\frac{1}{{S_k}^{(x+1)}}\sum_{i,j \in \Omega} {\gamma_{ijk}}^{(x)}(r^*_{ij}-(u_i^T)v_j)^2,
   \end{equation}
where $S$ is the total number of non-missing data points, $S_k^{(x+1)}$ is the sum of $\gamma_{ijk}$ for $k$-th Gaussian component and $x$ is the total number of iterations EM algorithm runs until convergence. Then we update the model parameters $U$ and $V$. We can rewrite the portion in Eq.  (\ref{eq:maxlikerewritten}) which is related to $U$ and $V$ as:

\begin{align} \label{eq:L2LRMF}
    \max_{V,U} & \sum_{i,j \in \Omega} \sum_{k=1}^{K} \gamma_{ijk}\bigg(-\frac{(r^*_{ij}-(u_i^T)v_j)^2}{2\sigma_k^2}\bigg) \nonumber \\ 
    = & -\sum_{i,j \in \Omega} \bigg(\sum_{k=1}^{K}\frac{\gamma_{ijk}}{2\sigma_k^2} \bigg) (r^*_{ij}-u_i^Tv_j)^2 \nonumber \\ 
    = & -\sum_{i,j \in \Omega} w_{ij}  (r^*_{ij}-u_i^Tv_j)^2  ,
\end{align}

\noindent where $w_{ij}$ represents the weight for each true rating $r_{ij}$, given by:
\begin{equation*}
   w_{ij} = \begin{cases}
    \sqrt{\sum_{k=1}^{K}\frac{\gamma_{ijk}}{2\sigma_k^2}}  ,& \text{if } i,j \in \Omega\\
    0,              & \text{if } i,j \notin \Omega.
\end{cases}
\end{equation*}
Eq. (\ref{eq:L2LRMF}) is equivalent to a weighted low-rank MF problem as given below:
\begin{equation*}
     \min_{U,V} W \odot (X-UV^T)^2.
\end{equation*}

The weighted low rank MF problems can be solved using methods such as Weighted Low-Rank Approximation \cite{srebro2003}, Damped Newton \cite{buchanan2005} and Weighted PCA \cite{torre2003}. We use Weighted PCA in this work to re-estimate model parameters $U$ and $V$. The EM algorithm stops alternating between E-step and M-step when two consecutive user latent factor matrices $U$ cause a change smaller than the given threshold value or the number of iterations reaches the pre-defined threshold. Algorithm 2 details the process of how MoG with MF model estimates noise and predict missing ratings.

\begin{algorithm}[ht]
\caption{Noise Estimation and Rating Prediction Model}
\begin{algorithmic}[1]
    \State{\textbf{Input: Perturbed Ratings ($R^*$)}}
    \State{\textbf{Output: $U$ and $V$}}
    \State \textit{Initialization}: Model parameters $U,V,\Pi$ and $\Sigma$ are randomly initialized
    \State In E-step posterior responsibility ${\gamma_{ijk}}^{(x)}$ is estimated using Eq. (\ref{eq:posteriodresp})
    \State {\textbf{For} {Until convergence}}
    \State $\> \>$ (M-Step for updating $\Sigma^{(x+1)}$ and $\Pi^{(x+1)}$) Model parameters $\Sigma$ and $\Pi$ are computed using Eq. (\ref{eq:paramupdate})
    \State $\>$ (M-Step for estimating $V$ and $U$) Model parameters $U$ and $V$ are updated using Eq. (\ref{eq:L2LRMF}) 
    \State $\>$ (E-step for posterior responsibility $\gamma_{ijk}$) posterior responsibility $\gamma_{ijk}$ is computed using current model parameters
    \State \textbf{Return} User and Item latent factor matrices $U$ and $V$
\end{algorithmic} 
\end{algorithm}

\section{Experimental Evaluation}
In this section, we evaluate the effectiveness of our proposed recommendation model through real-world datasets.
\subsection{Datasets}
We use three datasets: Movielens, Libimseti and Jester in the evaluation. Table \ref{tab:rating} provides a detailed view of the datasets. For privacy budget $\varepsilon$, we consider the value range from 0.1 to 3, lower values of privacy budget $\varepsilon$ guarantee stronger privacy protection for users.

\begin{table}[ht]
\caption{Rating Datasets}
\begin{center}
 \begin{tabular}{|p{35pt}|p{40pt}|p{30pt}||p{30pt}||p{40pt}|} 
 \hline
 Dataset & Total Ratings & No of Items & No of Users & Rating Scale  \\ [0.5ex] 
 \hline\hline
 Movielens & 100k & 1682 & 943 & 0.5 to 5  \\ 
 \hline
 Jester  & 2 Million & 100 & 73,421 & -10 to 10 \\
 \hline
 Libimseti  & 17,359,346 & 168,791  &  135,359 & 1 to 10 \\
 \hline
\end{tabular}
\end{center} \label{tab:rating}
\vspace{-6mm}
\end{table}
\subsection{Evaluation Metrics}
We use the Root Mean Squared Error (RMSE) and F-score to evaluate the recommendation system accuracy. We calculate the RMSE values over 10-fold cross-validation. RMSE can be estimated as :
\begin{equation*}
RMSE=\sqrt{\frac{\sum_{i=0}^{n-1}{(r_i-\hat{r_i})^2}}{n}},    
\end{equation*}
where $r_i$ is the actual rating, $\hat{r_i}$ is the predicted rating and $n$ is the total number of ratings in the aggregated dataset. In addition to RMSE, we use F-score to evaluate the performance of our recommendation system. Table \ref{tab:confusionmatrix} provides a detailed visualisation of how good a recommendation model is at predicting recommendations. Positives represent recommended items and negatives represent non-recommended items.

\begin{table}[ht] 
\caption{Confusion Matrix}
\begin{tabular}{|l|l|l|l|}
\hline
 & \multicolumn{3}{c|}{Actual Recommendations} \\ \hline
 \multicolumn{1}{|c|}{} 
&           & Positives      & Negatives      \\[1ex]\cline{2-4}
\begin{tabular}[c]{@{}l@{}}Predicted\\ Recommendations\end{tabular}
& Positives & True Positive  & False Positive \\ \cline{2-4} 
& Negatives & False Negative & True Negative \\[1ex]\hline
\end{tabular}\label{tab:confusionmatrix}
\end{table}

\noindent Precision and recall can be computed as follow:
 \begin{equation*}
     precision = \frac{\#true \ positives}{\#true \ positives \ + \ \#false \ positives},
 \end{equation*}
  \begin{equation*}
     recall = \frac{\#true \ positives}{\#true \ positives \ + \ \#false \ negatives}.
 \end{equation*}

\noindent We calculate F-score over top-10 recommended items. F-score can be computed as :
\begin{equation*}
    F \textnormal{-}Score = 2* \frac{Recall*Precision}{Recall+Precision}.
\end{equation*}

\subsection{Results}
\subsubsection{Noise Distribution Evaluation}
\begin{figure}[ht]
     \centering
        \subfloat[$\varepsilon=1$]{\label{fig:BLPMovielensdist1} \includegraphics[width=0.45 \textwidth]{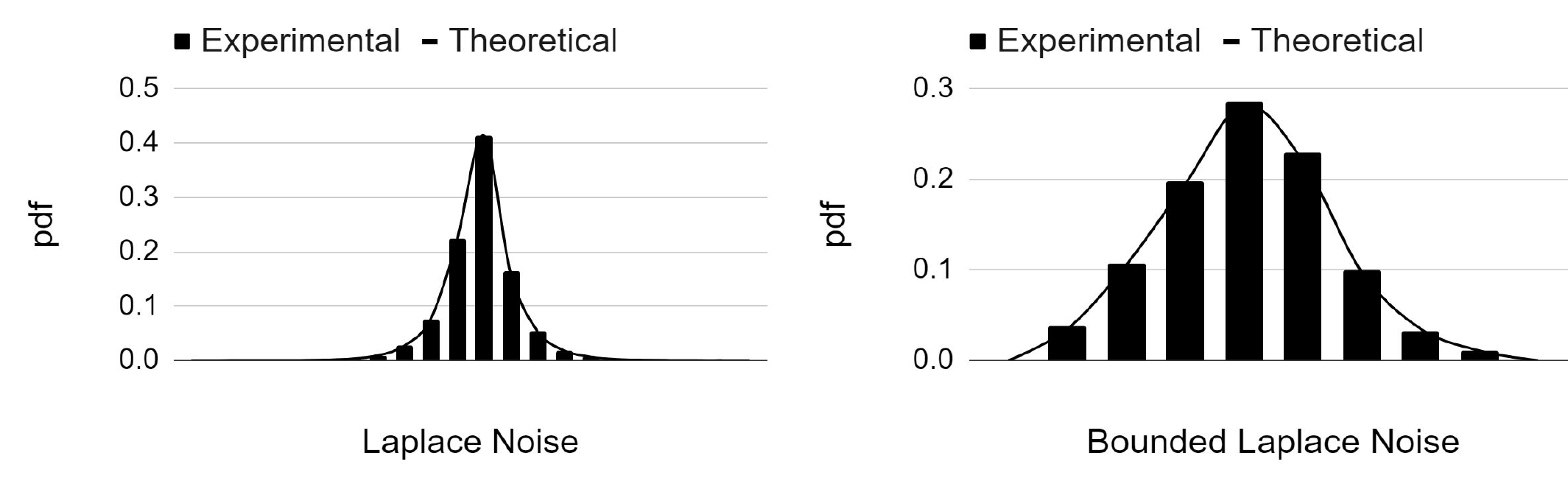}} \\
          \subfloat[$\varepsilon=0.1$]{\label{fig:BLPMovielensdist01}\includegraphics[width=0.45 \textwidth]{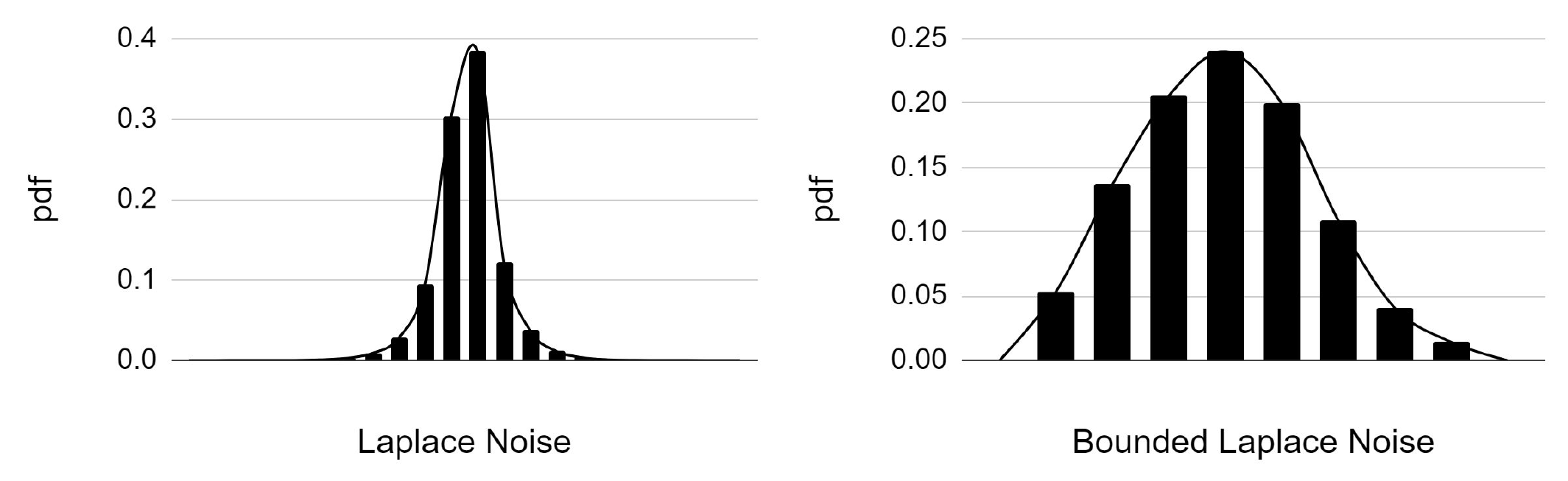}}
     \caption{Laplace vs Bounded Laplace Noise Distribution}
     \label{fig:noisedistribution}
\end{figure}

We derive the BLP noise distribution theoretically in section \ref{subsec:noisedistribution}. In this section, we show that the noise distribution of Laplace and BLP mechanisms are distinct. We generate 100,000 random noise samples using BLP and Laplace mechanisms for Movielens dataset while positioning their privacy budget $\varepsilon$ to 0.1 and 1. Fig. \ref{fig:BLPMovielensdist1} and \ref{fig:BLPMovielensdist01} display the probability of noise samples drawn by Laplace and Bounded Laplace mechanisms. From the probability density functions, we note that the noise distribution of the two mechanisms is distinct. We also plot the BLP noise distribution curve based on our derived noise distribution expressions given by Eq. (\ref{NoiseDistribution1}) and (\ref{NoiseDistribution2}). Fig. \ref{fig:BLPMovielensdist1} and \ref{fig:BLPMovielensdist01} show that the theoretical derivation of distribution follows the experimental distributions exactly.  

\subsubsection{Influence of BLP on predictive accuracy} \label{subsec:influenceBLP}
\begin{figure}[ht]
     \centering
     \subfloat[Movielens ]{\label{fig:BLPMovielens} \includegraphics[width=0.45 \textwidth]{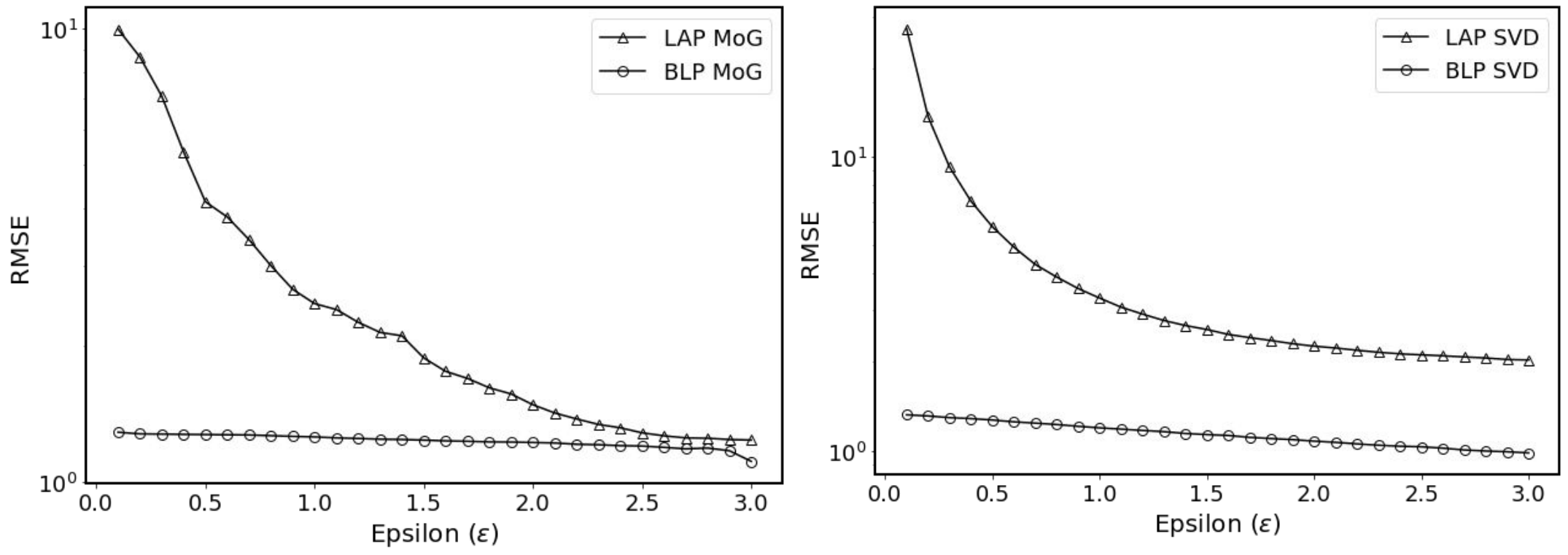}} \\
     \subfloat[Jester ]{\label{fig:JesterBLP} \includegraphics[width=0.45 \textwidth]{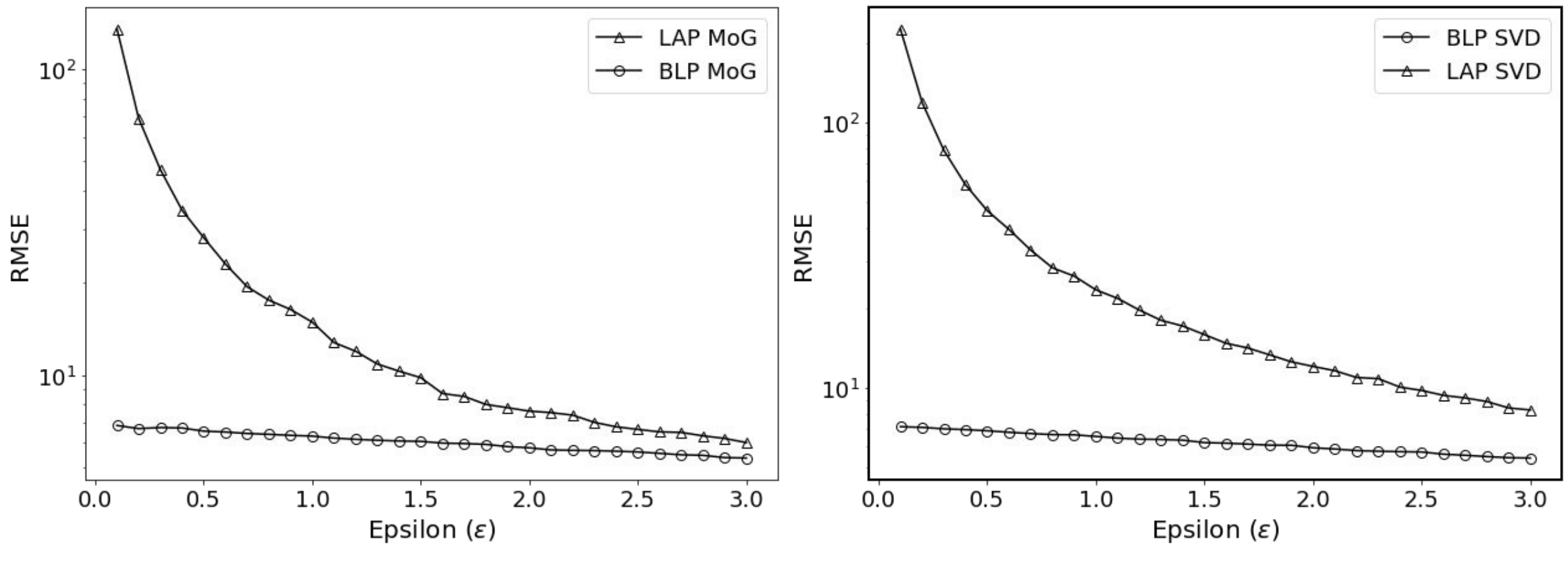}}
     \caption{Bounded Laplace Mechanism vs Laplace Mechanism RMSE Comparison}
     \label{fig:BLPLAPMovielens}
\end{figure}

In this experiment, we demonstrate that using BLP as input perturbation mechanism does play a significant role in obtaining higher predictive accuracy. We measure the RMSE when using either BLP or Laplace as the input perturbation mechanism while using the same rating prediction model (either MoG or SVD). Fig. \ref{fig:BLPMovielens} and \ref{fig:JesterBLP} display the resulting RMSE metric values for Movielens and Jester datasets respectively. For both datasets and rating prediction models, BLP mechanism results in higher predictive accuracy than the Laplace mechanism.
\subsubsection{Influence of MoG on predictive accuracy} \label{subse:MoGaccuracy}
\begin{figure}[ht]
     \centering
     \subfloat[Movielens]{\label{fig:MovielensGMM} \includegraphics[width=0.45 \textwidth]{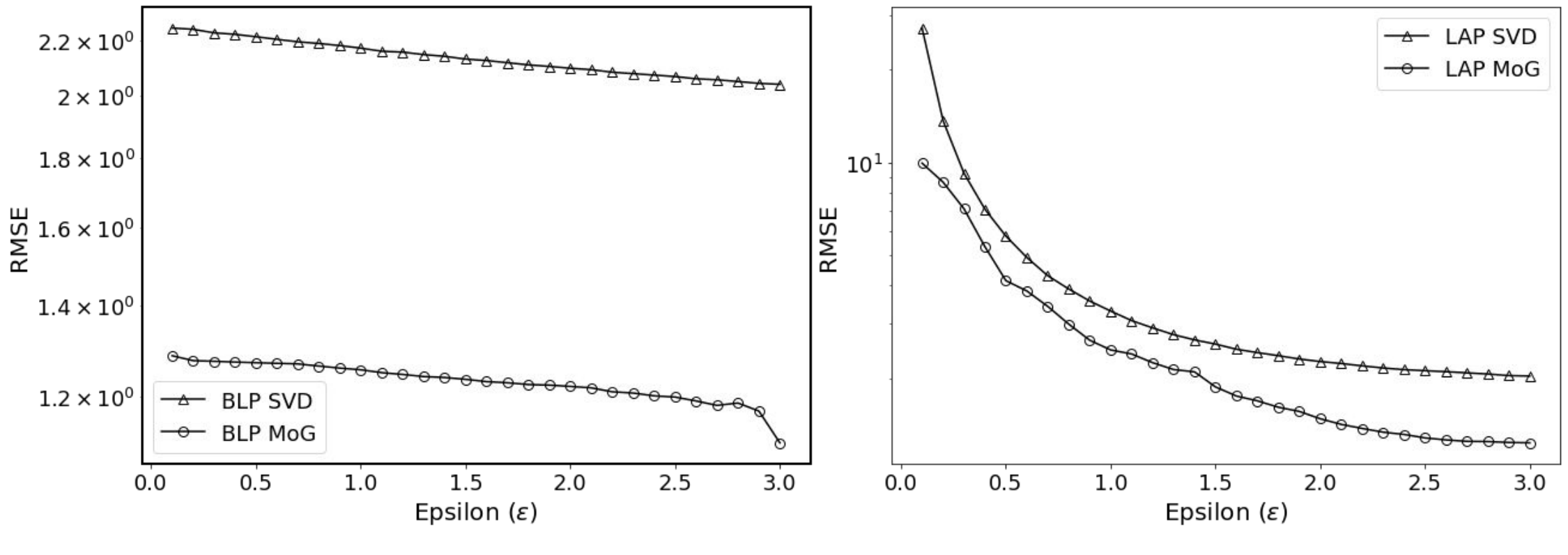}} \\
     \subfloat[Jester]{\label{fig:JesterGMM} \includegraphics[width=0.45 \textwidth]{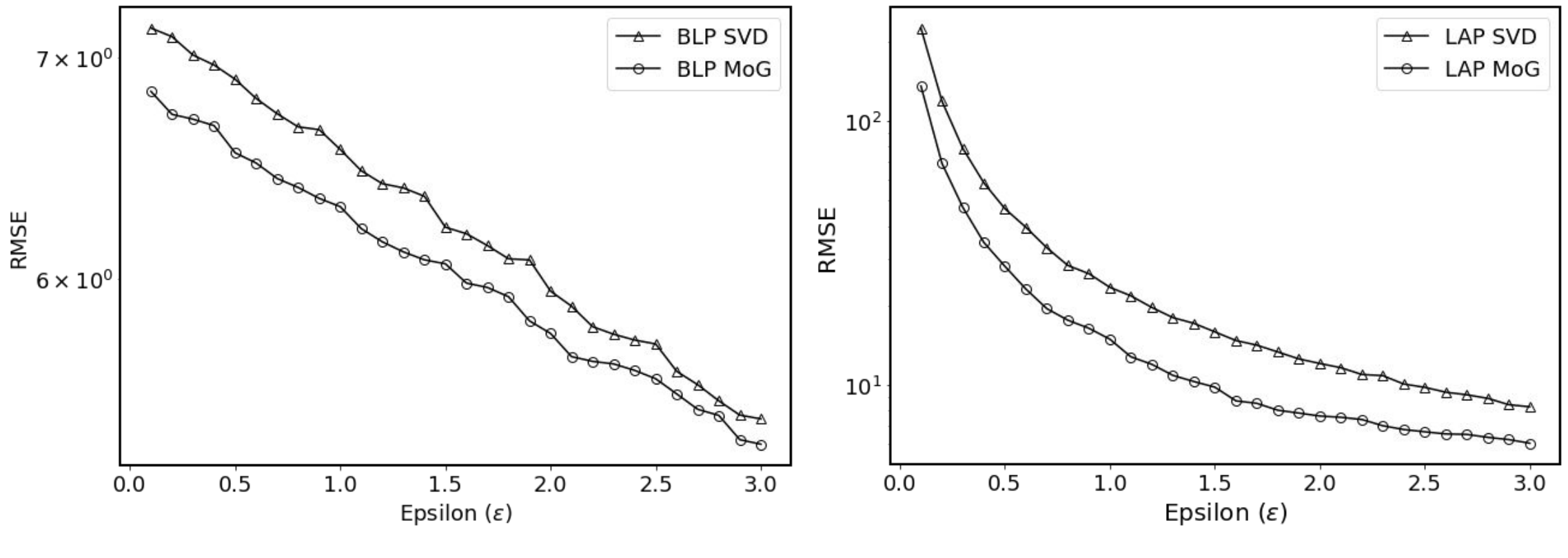}}
     \caption{MoG vs SVD Prediction Model RMSE Comparison}
\end{figure}

In this experiment, we demonstrate that employing MoG in our recommendation model aids to improve predictive accuracy for lower values of privacy budget $\varepsilon$. We measure the RMSE values when using the MF with MoG or the SVD for rating prediction while using the same data perturbation mechanism. Fig. \ref{fig:MovielensGMM} and \ref{fig:JesterGMM} display the resulting RMSE values for Movielens and Jester datasets respectively. For both datasets, the predictive accuracy from the MoG prediction model is much higher than SVD.

\subsubsection{Predictive accuracy comparison over other private recommendation models} \label{subsec:comparisonprediction}
We compare the predictive accuracy of our recommendation model with other existing local differentially private recommendation models such as:
\begin{itemize}
    \item Input Perturbation Method (ISGD) \cite{berlioz2015}: This method perturbs the user's original ratings locally using the Laplace mechanism. However, they apply a truncation method to ensure that the perturbed rating falls within a pre-defined domain. The noised ratings which fall out of a pre-defined range are clamped to either lower or upper bound of the rating domain using a threshold value. ISGD method uses MF for rating prediction at SP side.
    \item Private Gradient-Matrix Factorization (PG-MF) \cite{shin2018}: This approach uses MF to perform recommendations. In this approach, user computes user latent factors locally without submitting them to the SP. The SP estimates the item latent factors after collecting gradients from the users. Users, on the other hand, compute a perturbed gradient and submit that to the SP. The SP aggregates the perturbed gradient from all the users and then update the item latent factors accordingly. This method requires iterative communication between users and SP.
\end{itemize}
We use Non-Private MF as the baseline method as it does not use any local perturbation mechanism to perturb user's original ratings. Instead, MF algorithm uses original ratings to predict missing ratings. The baseline method provides us with a lower bound RMSE value for predictive error. Our recommendation model (BLP-MoG-MF) uses BLP as input perturbation mechanism and MoG-MF as recommendation algorithm. BLP-MoG-MF method uses objective function specified by Eq. (\ref{eq:Objectivefunction}) to obtain latent factor matrices. ISGD and PG-MF methods also use the same objective function to perform rating predictions. To maintain the fairness of comparison, we did not compare our results with recommendation models which use different approaches to predict missing ratings.  

\begin{figure}[ht]
\includegraphics[width=1 \linewidth]{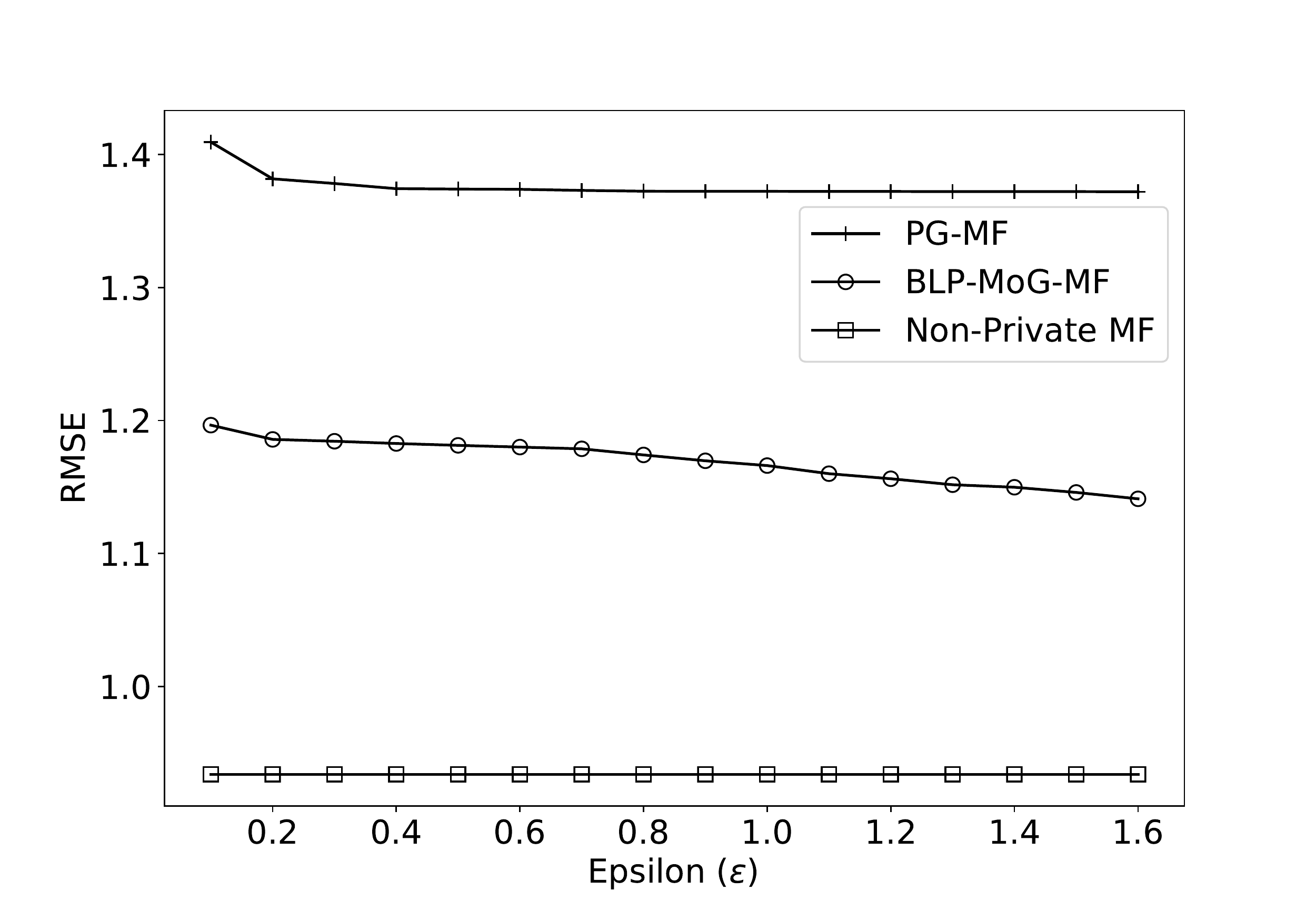} \centering
\caption{PG-MF vs BLP-MoG-MF RMSE Comparison for Movielens}
\label{fig:ComparisionMovielens}
\end{figure}

\begin{figure}[ht]
\includegraphics[width=1 \linewidth]{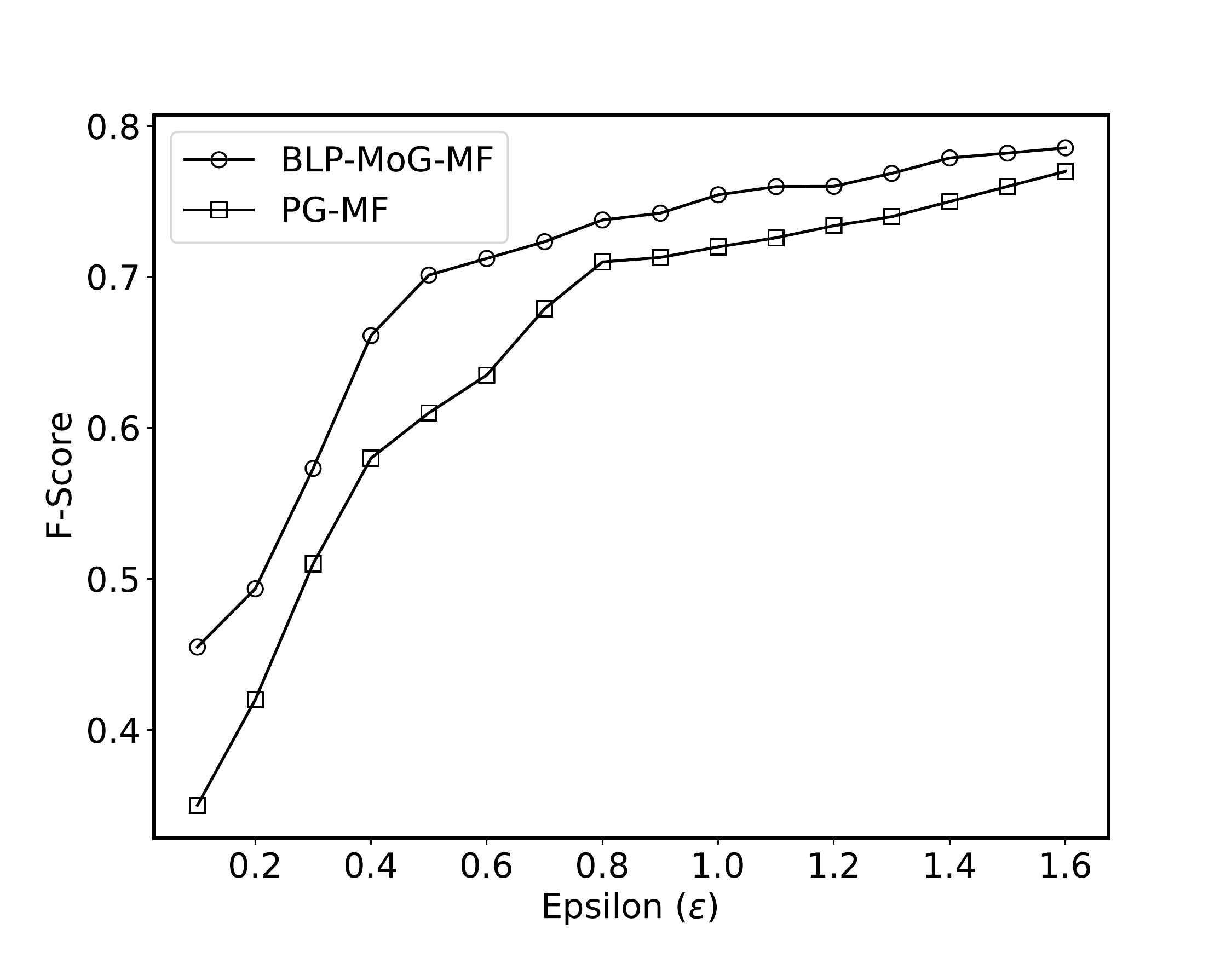} \centering
\caption{PG-MF vs BLP-MoG-MF F-Score Comparison for Movielens}
\label{fig:ComparisionMovielensfscore}
\end{figure}

Firstly, we compare BLP-MoG-MF with PG-MF. We vary the privacy budget $\varepsilon$ from 0.1 to 1.6 for Movielens dataset. Fig. \ref{fig:ComparisionMovielens} displays the RMSE values for BLP-MoG-MF, PG-MF and the baseline method. As expected, when the privacy budget increases, predictive accuracy for all privacy protection methods increases. Because when privacy budget $\varepsilon$ increases, the magnitude of privacy loss input perturbation mechanism permits increases, which in succession, causes a rise in the predictive accuracy. More importantly, we notice that BLP-MoG-MF provides a lower RMSE than PG-MF for the same privacy budget $\varepsilon$. Fig. \ref{fig:ComparisionMovielensfscore} demonstrates the F-score values for BLP-MoG-MF and PG-MF method. Similar to RMSE values, F-score values also increases when the privacy budget $\varepsilon$ increases. F-score values indicate that BLP-MoG-MF provides more accurate recommendations compared to PG-MF for all values of privacy budget $\varepsilon$.

\begin{figure}
     \centering
     \subfloat[Movielens]{\label{fig:isgd1} \includegraphics[width=0.8 \linewidth]{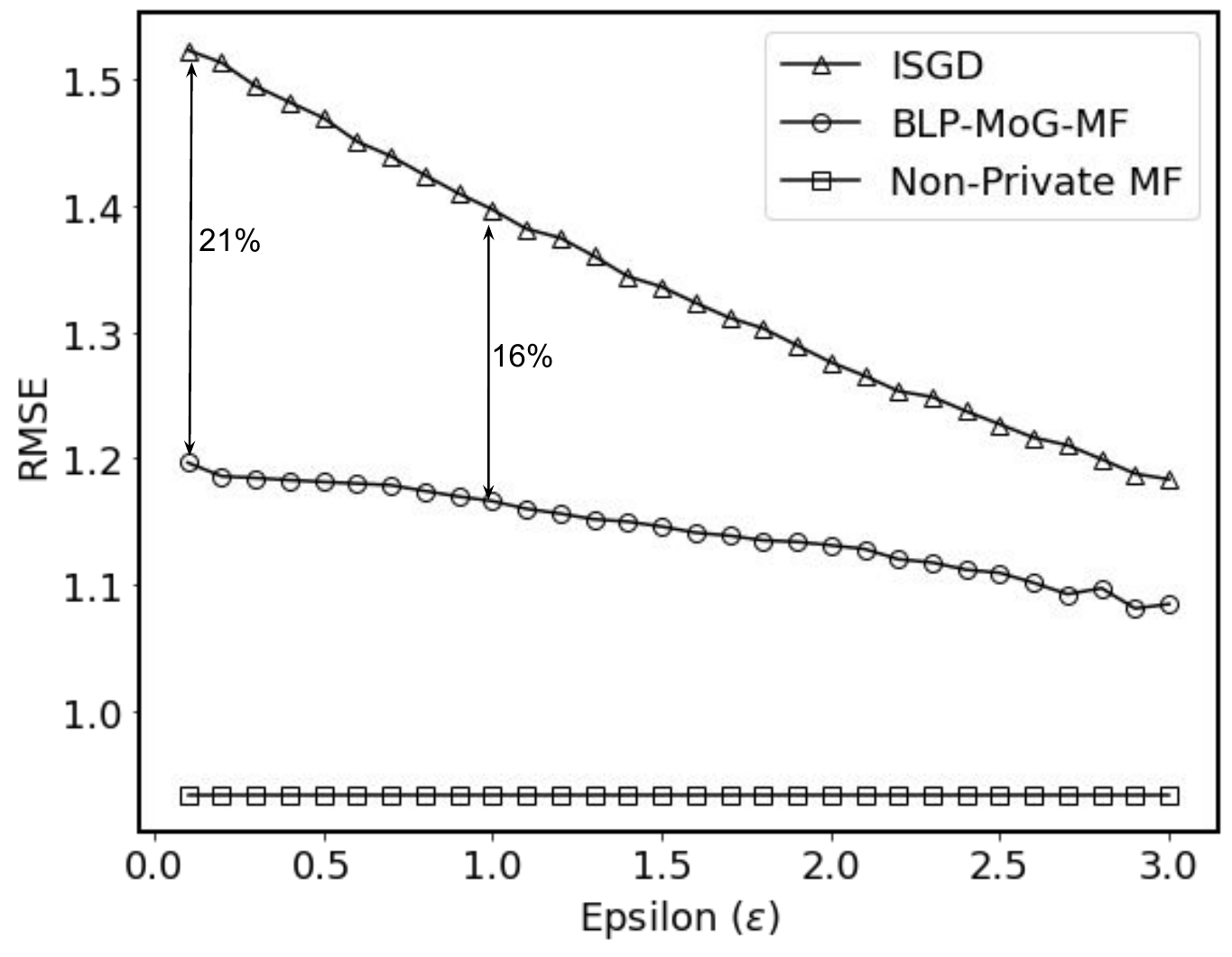}} \\
     \subfloat[Jester]{\label{fig:isgd2} \includegraphics[width=0.8 \linewidth]{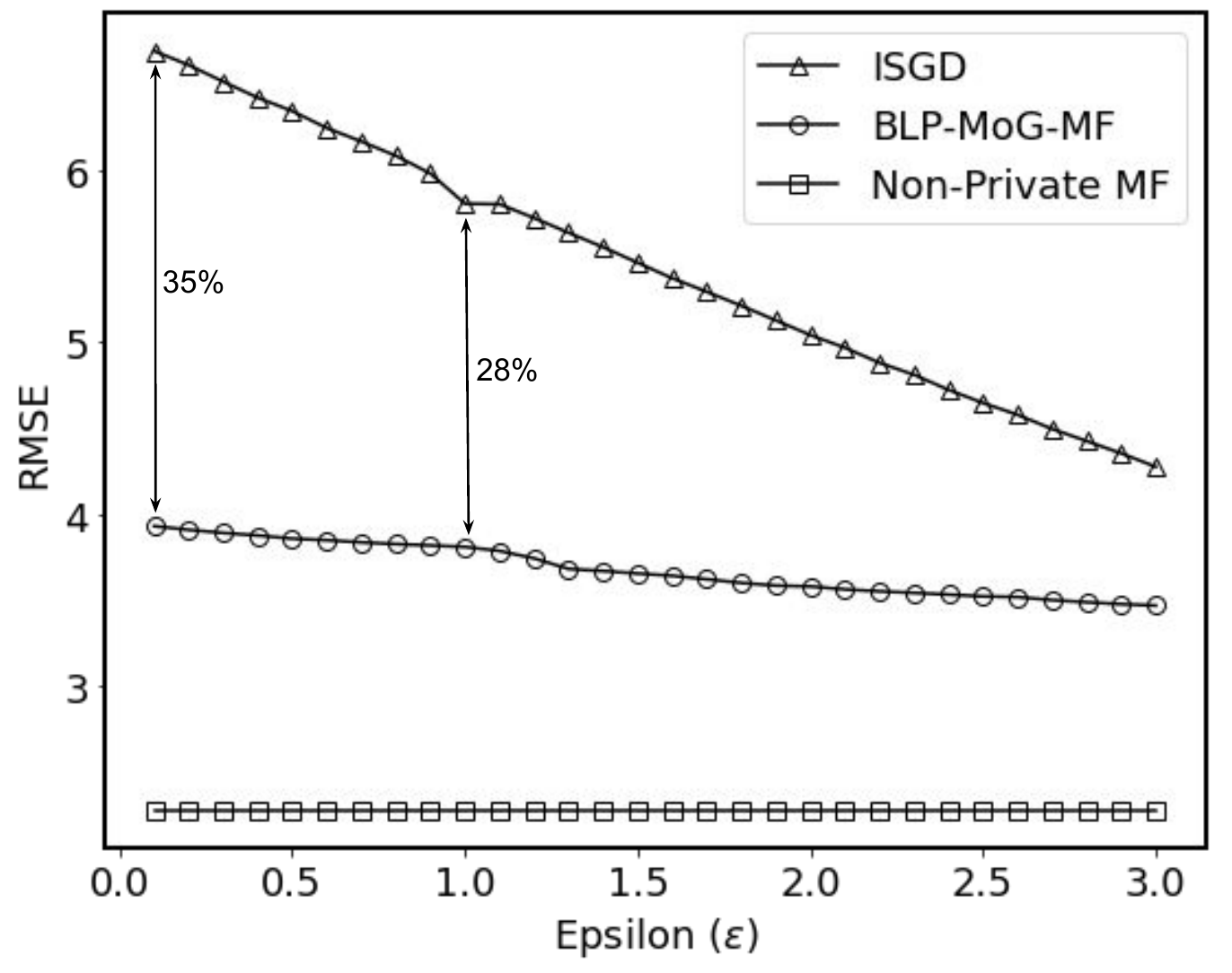}} \\
     \subfloat[LibimSeti]{\label{fig:isgd3} \includegraphics[width=0.8 \linewidth]{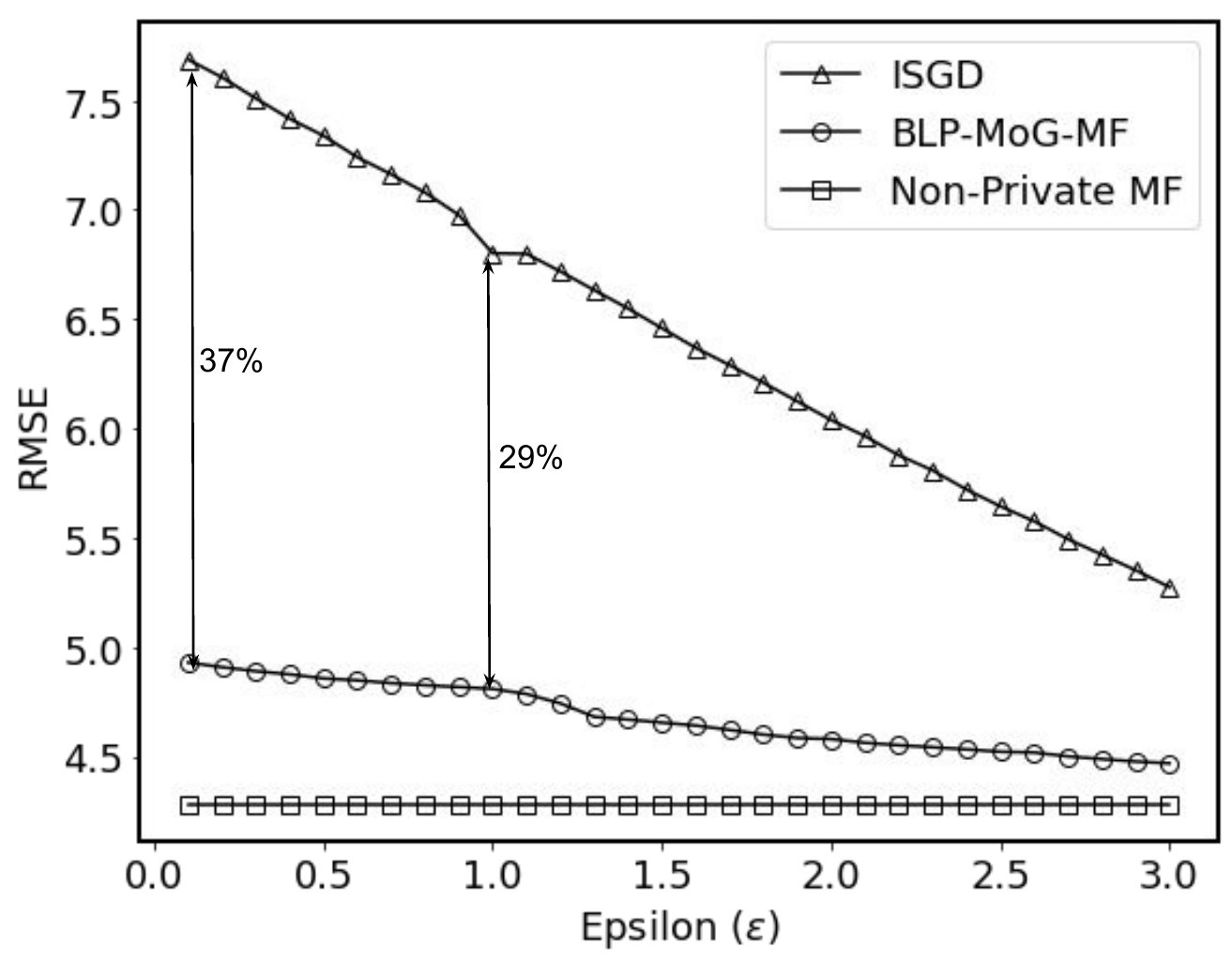}} \\
     \caption{BLP-MoG-MF vs ISGD RMSE Comparison}
\end{figure}

\begin{figure}
     \centering
       \subfloat[Movielens]{\label{fig:Movielensfscore} \includegraphics[width=0.8 \linewidth]{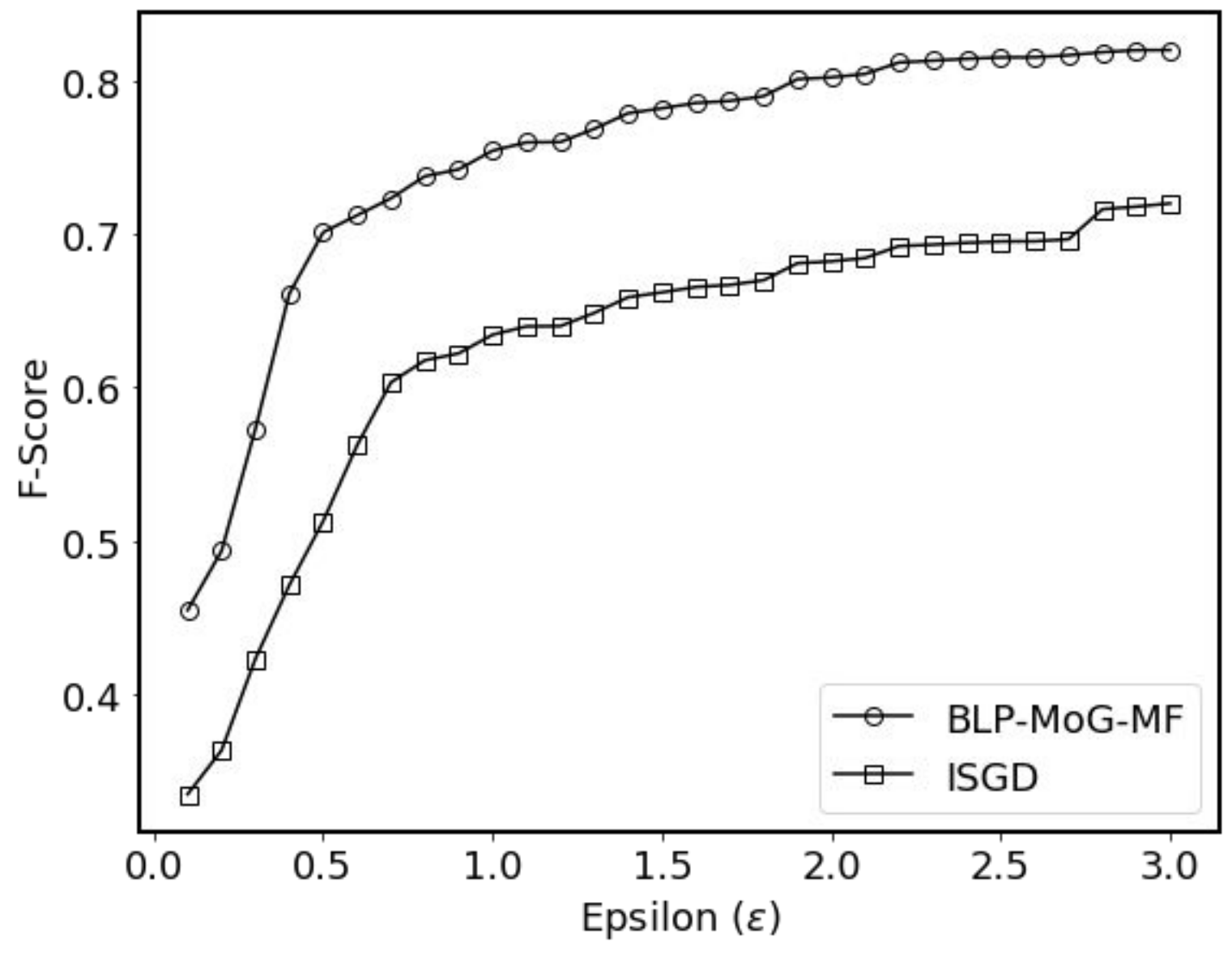}} \\
     \subfloat[Jester]{\label{fig:Jesterfscore} \includegraphics[width=0.8 \linewidth]{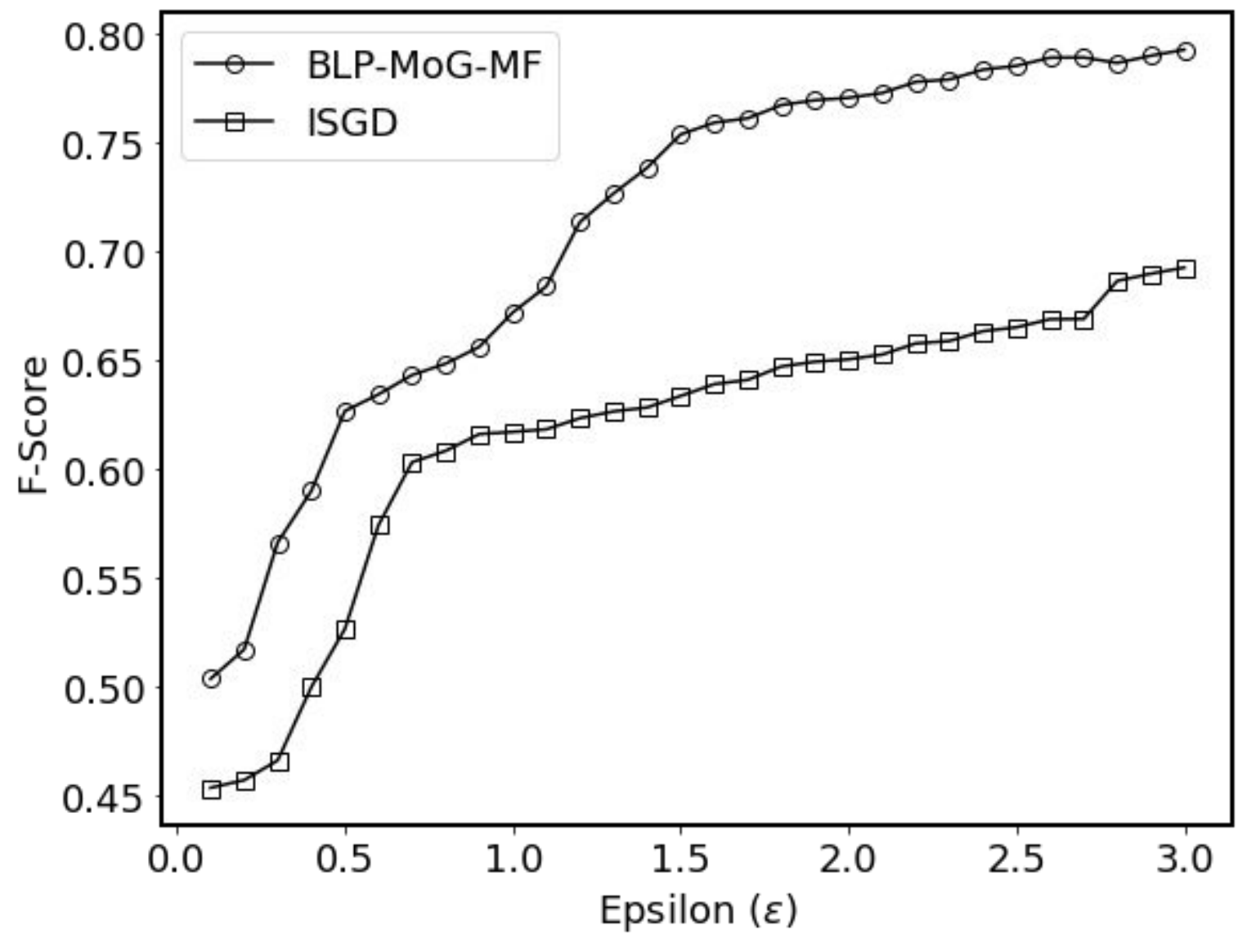}} \\
     \subfloat[LibimSeti]{\label{fig:libimfscore} \includegraphics[width=0.8 \linewidth]{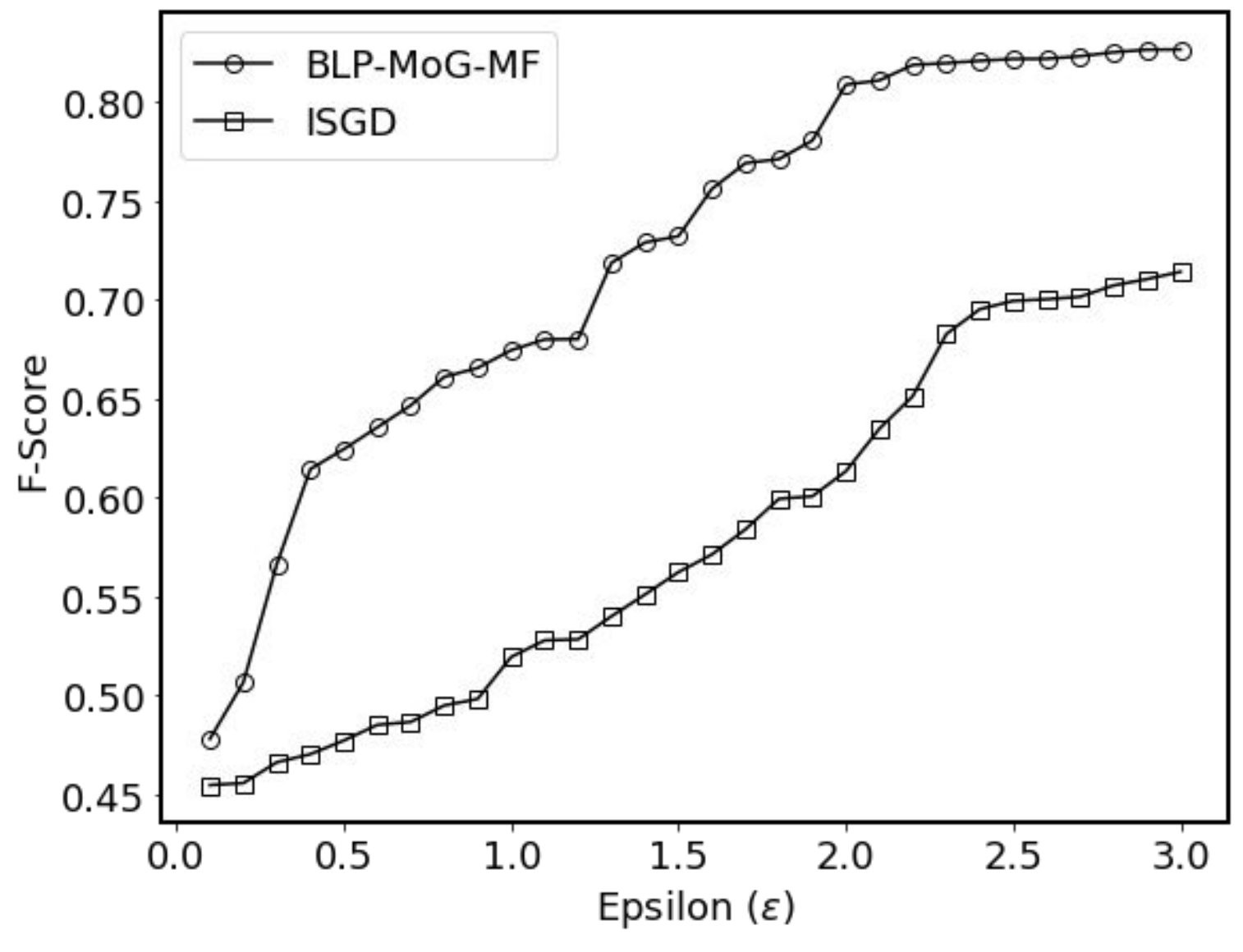}} \\
    \caption{BLP-MoG-MF vs ISGD RMSE Comparison}
\end{figure}

Then, we compare BLP-MoG-MF with ISGD method for Movielens, Libimseti and Jester datasets. We vary the privacy budget $\varepsilon$ from 0.1 to 3 for all the datasets in this simulation. Fig. \ref{fig:isgd1}, \ref{fig:isgd2} and \ref{fig:isgd3} display the RMSE values for BLP-MoG-MF, ISGD and the baseline methods. The results show that BLP-MoG-MF outperforms ISGD significantly for all values of the privacy budget $\varepsilon$. Fig. \ref{fig:Movielensfscore}, \ref{fig:Jesterfscore} and \ref{fig:libimfscore} illustrate the F-score values for Movielens, Jester and LibimSeti datasets for both BLP-MoG-MF and ISGD method. We vary the privacy budget $\varepsilon$ from 0.1 to 3 for all the datasets and both methods. The F-score gets larger as privacy budget $\varepsilon$ increases for all the datasets and both methods. Additionally, for all datasets, the F-scores of BLP-MoG-MF method is higher than the ISGD method. This trend implies that our method guarantees higher data utility for all the values of privacy budget $\varepsilon$.

\subsubsection{Analysis of Communications Cost} \label{subsec:communicationcost}
We compare the communication cost incurred in our approach at each iteration to recommendation models proposed by \cite{shin2018} and \cite{berlioz2015}. Table \ref{tab:commcost} summarises the analysis. Both BLP-MoG-MF and ISGD methods require the user to transfer a perturbed rating whenever user rates an item. In PG-MF method, the user transmits the perturbed gradient of user-latent factors to SP over multiple data transmission iterations. Both BLP-MoG-MF and ISGD methods do not require the SP to transmit any data back to the user. However, in PG-MF approach at each iteration, the SP transmits an updated item latent factor matrix back to the user. This exchange between the SP and the user continues until the number of iterations reaches a pre-defined threshold value. We assume a single rating is 1 bit. The estimated size of the transmitted data for each iteration for PG-MF method is approximately 0.15 MB for Movielens dataset [1].
\begin{table}
\caption{Comparison of Communication Cost for Movielens}
\begin{center}
 \begin{tabular}{|p{70pt}|p{60pt}|p{60pt}|} 
 \hline
 Recommendation Model & User to SP & SP to User  \\ [0.5ex] 
 \hline\hline
 BLP-MoG-MF & 1 bit & No transfer  \\ 
 \hline
 ISGD  & 1 bit & No transfer \\
 \hline
 PG-MF  & 1 bit & 0.15MB  \\
 \hline
\end{tabular}
\end{center} \label{tab:commcost}
\end{table}

This shows that we significantly reduce the communication cost in our proposed model compared to other local differential private recommendation models.

\section{Conclusion}
In our work,  we have proposed a recommendation model under the consideration of an untrustworthy service provider. We have used BLP as local input perturbation mechanism and MoG-MF for noise estimation and rating prediction. Compared to existing solutions, our proposed recommendation model can improve predictive accuracy and guarantees strong user privacy. Besides, our method does not incur any further communication cost to the user side as it only requires the user to transmit the perturbed rating to the SP.

\appendices \label{app:prooftheorem}
\section{Proof of Lemma 2}
Assume $r$ and $r'$ are a pair of possible inputs of the BLP mechanism where $r' \geq r$ and $r'=r+z$. Let $0 \leq z \leq \Delta f$. In order to prove Lemma 2, we must first consider few other properties concerning $C(r)$. First we find $\frac{\partial}{\partial z} F(r,z) \geq 0 $ when $r+z \leq u$.

\begin{equation*}
\begin{split}
   & \frac{\partial}{\partial z} F(r,z) = \frac{1}{C(r)}\frac{\partial}{\partial z} \bigg(C({r+z}) e^{\frac{z}{b}} \bigg) \\
    &= \frac{1}{C(r)}\frac{\partial}{\partial z} \bigg (e^{\frac{z}{b}} - \frac{1}{2} \big(e^{-\frac{(r+z)-l}{b}} - e^ {-\frac{u-(r+z)}{b}}) e^{\frac{z}{b}} \bigg)  \\
    &=\frac{1}{C(r)}\frac{\partial}{\partial z} \bigg (e^{\frac{z}{b}} - \frac{1}{2}(e^{\frac{
    -r+l}{b}} - e^ {\frac{-u+r+2z}{b}})  \bigg) \\
    &=\frac{1}{C(r)b} \bigg(1- e^ {-\frac{u-r-z}{b}} \bigg) e^{\frac{z}{b}}
\end{split}
\end{equation*}
As $b > 0$, we then see that $\frac{\partial}{\partial z} F(r,z) \geq 0 $ when $r+z \leq u$. 

\noindent Then we prove that $\frac{\partial}{\partial r} F(r,z) \leq 0 $ when $z \geq 0$. First we note, 
\begin{equation*}
    \frac{\partial}{\partial r}C({r+z}) = \frac{1}{2b} \bigg( e^{-\frac{-r+z-l}{b}} - e^ {-\frac{u-r-z}{b}} \bigg)
\end{equation*}
We find,
\begin{equation*}
\begin{split}
    &\frac{\partial}{\partial r} F(r,z) \\
    &= \frac{e^\frac{z}{b}}{C(r)^2} \bigg( C(r) \frac{\partial}{\partial r}C({r+z}) - C({r+z})\frac{\partial}{\partial r}C({r}) \bigg) \\
    &=\frac{e^\frac{z}{b}}{2bC({r})^2} \bigg(e^{-\frac{r-l}{b}}(e^{-\frac{z}{b}}-1) +e^{-\frac{u-l-z}{b}} 
    +e^{-\frac{u-r}{b}}(1-e^{\frac{z}{b}}) \\
    &- e^{-\frac{u-l+z}{b}} \bigg) \\
    &=\frac{e^{\frac{z}{b}} \bigg( \bigg( e^{-\frac{z}{b}}-1 \bigg) \bigg( e^{\frac{u-r}{b}}-1 \bigg)+ \bigg( 1-e^{\frac{z}{b}} \bigg) \bigg( e^{\frac{r-l}{b}}-1 \bigg) \bigg)}{2be^{\frac{u-l}{b}}C(r)^2}
\end{split}
\end{equation*}
Since $r \in [l,u]$, it proves that $e^{\frac{u-r}{b}},e^{\frac{r-l}{b}} > 1$. When $z \geq 0$, it shows that $e^{-\frac{z}{b}} < 1$ and $e^{\frac{z}{b}}>1$. Therefore, $\frac{\partial}{\partial r} F(r,z) \leq 0$ when $z \geq 0$.

As $\frac{\partial}{\partial r} F(r,z) \leq 0$, the maximum value of $F(r,z)$ at a fixed $z^0$ is attained at the smallest possible value of $r$, i.e $r=l$.

\begin{equation*}
    \max_{\substack{r,r+z^0 \in [l,u] \\ 0 \leq z^0 \leq \Delta f}} F(r,z^0)=\max_{0 \leq z^0 \leq \Delta f} \frac{C(l+z^0)}{C(l)}e^{\frac{z^0}{b}}
\end{equation*}

Then, as $\frac{\partial}{\partial z} F(l,z) \geq 0$, the maximum value of $F(l,z)$ is attained at the largest possible $z$, i.e $z=\Delta f$,

\begin{equation*}
\begin{split}
    \max_{0 \leq z \leq \Delta f} \frac{C(l+z)}{C(l)}e^{\frac{z}{b}} &= \frac{C(l+\Delta f)}{C(l)}e^{\frac{\Delta f}{b}} 
\end{split}
\end{equation*}

% use section* for acknowledgment

% Can use something like this to put references on a page
% by themselves when using endfloat and the captionsoff option.
\ifCLASSOPTIONcaptionsoff
  \newpage
\fi

\end{document}